\documentclass[10pt,twocolumn,letterpaper]{article}

\usepackage{wacv}
\usepackage{times}
\usepackage{epsfig}
\usepackage{graphicx}
\usepackage{amsmath}
\usepackage{amssymb}

\usepackage[pagebackref=true,breaklinks=true,letterpaper=true,colorlinks,bookmarks=false]{hyperref}

\wacvfinalcopy 

\def\wacvPaperID{****} %

\ifwacvfinal\fi
\begin{document}

\title{Deep Convolutional Features for Image Based Retrieval and Scene Categorization}

\author{Arsalan Mousavian\\
George Mason University\\
4400 University Dr Fairfax, VA, 22032\\
{\tt\small amousavi@gmu.edu}
\and 
Jana Ko\v{s}eck\'a\\
George Mason University\\
4400 University Dr Fairfax, VA, 22032\\
{\tt\small kosecka@gmu.edu}
}

\maketitle

\begin{abstract}
   Several recent approaches showed how the representations learned by Convolutional Neural Networks can be repurposed 
   for novel tasks. Most commonly it has been shown that  the activation features of the last fully connected layers (fc7 or fc6) 
   of the network, followed by a linear classifier outperform the state-of-the-art on several 
   recognition challenge datasets.  Instead of recognition, this paper focuses on the image retrieval problem
   and proposes a examines alternative pooling strategies derived for CNN features.
   The presented scheme uses the features maps from an earlier layer 5 of the CNN architecture, which has been shown to preserve coarse spatial information and is semantically meaningful. 
   We examine several pooling strategies and demonstrate superior performance 
   on the image retrieval  task (INRIA Holidays) at the fraction of the computational cost, while using a relatively small 
   memory requirements.  In addition to retrieval, we see similar efficiency gains on the SUN397 scene categorization   dataset, demonstrating wide applicability of this simple strategy.  We also introduce and evaluate a novel GeoPlaces5K dataset from different geographical locations in the world for image retrieval that stresses more dramatic changes in appearance and viewpoint.
\end{abstract}

\section{Introduction}

Past few years noted increased activity in the use of convolutional neural networks (CNN) for a variety 
of classical computer vision problems. The initial breakthroughs were enabled by the availability of large datasets 
(ImageNet, Places) yielding dramatic improvements on the object and scene classification task~\cite{Krizhevsky-NIPS2012}.
Since this initial success several strategies have been explored to adapt the 
network parameters or architecture to other tasks~\cite{DeCafDonahue-2013}.  Typical convolutional neural networks used for categorization tasks are often concatenations of multiple convolution and pooling layers followed by two or three  fully connected layers and a soft-max classifier.  
It has been demonstrated in \cite{Carlsson-CVPR2014} that using last fully connected layer features (fc7) 
from pre-trained CNNs \cite{Overfeat2013}  as a representation, is suitable for linear classifiers such as SVM, leads
to superior performance on a variety of classification tasks.  More comprehensive study of transferability of representations of features derived from CNN's to different tasks can be found in~\cite{Azizpour-arXiv14}. 

In this paper, instead of exploiting the features from fully connected layers  as image representation for the categorization and image retrieval tasks,  we propose significantly more efficient, compact, and more discriminant representation and associated pooling strategy. 
Using CNNs pre-trained on {\em Places} \cite{cnnPlaces-NIPS2014} and {\em ImageNet} ~\cite{Krizhevsky-NIPS2012} we consider the feature maps computed at the last pooling layer 5 before the fully connected layers.   We demonstrate that these features are more effective in retrieving instances of the same objects under dramatic variations of viewpoint and scale as encountered in INRIA Holiday dataset and show how different pooling strategies affect this capability.  More recently the effectiveness of max and average pooling strategies was also investigated in~\cite{Razavian-arXiv} in the context of image retrieval task. Related to the insights obtained previously, we propose additional hybrid pooling strategy, provide detail visualization of the effects of the pooling strategies and their dependence on clutter and viewpoint. This is supported by recent strategies for visualization of network layers as well as ablation studies presented in~\cite{ZeilerFergus-ECCV2014}. The intuition behind the effectiveness of our approach is that in the layers before last fully connected layers the encoded information is more semantically meaningful and spatially localized. At last we introduce and evaluate the retrieval accuracy on a new challenging GeoPlaces5K dataset containing images of different geographic locations taken at different times of day, with dramatic variations of viewpoints.  

\begin{figure*}
\centering
\includegraphics[width=\textwidth]{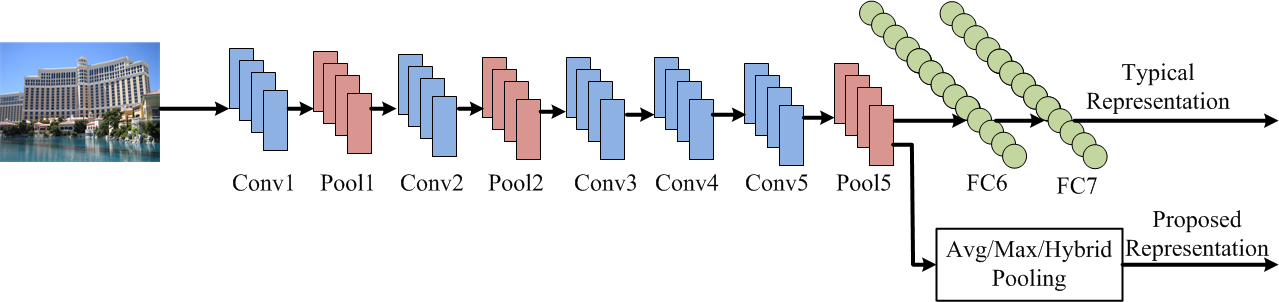}
\caption{Overview of Proposed Approach. Fully-connected layer 7 (fc7) of pre-trained networks on {\em ImageNet} or {\em Places} is commonly used as feature for retrieval/classification tasks. Our approach shows that earlier layer such as pool5 captures more general purpose semantics and is more suitable for general classification/retrieval application on the tasks related to the original training objective.  Furthermore it is not required to apply our method on multiple scales nor object proposals, which is desirable aspect for the efficiency. }
\label{fig:overview}
\end{figure*}

The overview of our method is shown in Figure~\ref{fig:overview}. 
In addition to the image retrieval task we also evaluate the proposed strategy on SUN397 scene categorization dataset achieving comparable performance to the state-of-the-art more efficiently and with order of magnitude smaller memory footprint. 
 

\section{Related Work}

Past few years have shown increased activity in the use of convolutional neural networks (CNN) for a variety 
of classical computer vision problems. The initial breakthroughs has been lead by improved accuracies 
on the image classification task~\cite{Krizhevsky-NIPS2012}  with CNN trained on the ImageNet object categorization dataset. 
Notable efforts were devoted to studies of effects of different modes of training and experimenting with different architectures~\cite{LeNg-ICML2012, Simonyan-NIPS2013} and \cite{DeCafDonahue-2013}. 
Since the initial success, CNN features~\cite{Overfeat2013} has been used as universal representation for a variety 
of classification tasks~\cite{Carlsson-CVPR2014} and \cite{DeCafDonahue-2013}. In addition to object categorization, the use of CNN architectures for object localization~\cite{Sivic-CVPR2014}, scene classification and other visual recognition tasks have been demonstrated. 
Attempts to use CNNs for semantic segmentation was lead by~\cite{DarrellSemSeg2014}.


Our approach is motivated by the efforts of understanding the representations learned by CNN's using visualization strategies, enabling both to observed learned invariances at different levels as well as tracing back high activations at the last fully connected layers back to image patches. These strategies provide some insight into factors which affect most the classification performance. 
In ~\cite{Zhou-ICLR2014}, authors demonstrated that dominant objects which contribute to scene classification, while 
in Zeiler et al \cite{ZeilerFergus-ECCV2014}showed that feature maps following the later convolutional layers  encode both spatial and semantic 
information of the dominant attributes and semantic concepts. 

Several works investigated the performance of CNN features with the goal of getting better understanding of the invariance properties as well as utility of the CNN representations for various classification tasks. Rigorous evaluation of the comparison of CNN methods with shallow representations such as Bag-of-Visual-Words and Improved Fisher vectors has been conducted in~\cite{Chatfield14return}. 
The evaluation was carried out on the different categorization tasks (ImageNet, Caltech and PASCAL-VOC). 
The premise of this study was to compare different representations which are suitable for the analysis with linear classier such as SVM. The experiments concluded that while the shallow methods can be improved using data augmentation, the CNN representations significantly improve the classification performance.
In the work of~\cite{Lazebnik-arXiv2014} authors proposed computation of  CNN features over windows at multiple scales and aggregating these representations in a manner similar to Spatial Pyramid Pooling, affecting favorably both the classification and image based retrieval performance. While the pooling strategy was found effective,  the features extraction stage was expensive, yielding high feature dimensionality. 
All the methods mentioned above used the last fully connected layer fc7 features as image or window representations with dimensionality of 4096. 
In the proposed work we argue for alternative CNN derived features and novel pooling strategy. Previously the convolutional level 5 features have been evaluated in the absence of pooling strategies on Caltech-101 dataset in~\cite{DeCafDonahue-2013}, yielding inferior performance compared to fully connected layer features fc6 and fc7 . 
With the exception of ~\cite{Lazebnik-arXiv2014} the above mentioned studies focus on classification instead of retrieval tasks. Another line 
of work is related to the image retrieval. Representations used in the past for the image-based retrieval used both local and global 
features. They often considered baseline method is the bag-of-visual-words representation, followed by spatial verification 
of top retrieved images using geometric constraints~\cite{Philbin-CVPR07}. Various improvements of these methods include learning better vocabularies, developing better quantization and spatial verification methods~\cite{ChumECCV10} or improving the scalability. Alternative more powerful quantization and representation techniques have been also explored in~\cite{Jegou-ICCV2012, Perronin-CVPR2012,Chum-CVPR2012}. The evaluation strategies of the image based retrieval strategies typically assume that the query instance is available in the reference dataset. The existing datasets vary in their size, the number of distractor images and the amount of clutter and viewpoint variation 
they exhibit. The most commonly used datasets INRIA Holidays~\cite{Jegou-ICCV2012}, Oxford Buildings~\cite{Philbin-CVPR07} and Kentucky dataset \cite{Nister-CVPR06}. 

Related image retrieval problem tackled in the past  is the problem of geo-location. The work of~\cite{Hays-CVPR08} proposed a data driven method for computing the coarse geographical location of an image using simpler features like GIST and color histograms. In this setting the exact instances of query views are often not available, but images in the reference set which share the same architectural style and appearance are likely to come from similar geographic locations. Some of these effects are evaluated and visualized on the new GeoPlaces dataset introduced in this paper and used to evaluate the retrieval accuracy. 
 \\

{\em 
}


\begin{figure}
\begin{tabular}{c@{\hspace{1.5mm}}c@{\hspace{1.5mm}}c@{\hspace{1.5mm}}c}
 & Tower  & Car/Roads & Color Blue \\
\includegraphics[width=0.11\textwidth , height=0.11\textwidth]{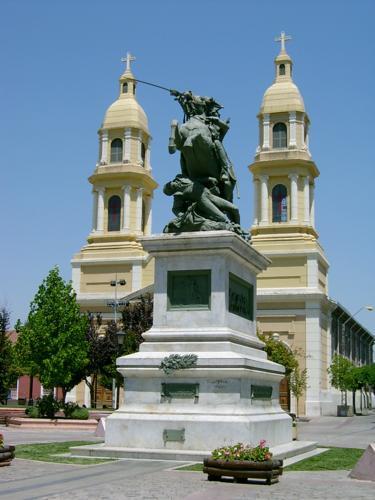}&
\includegraphics[width=0.11\textwidth]{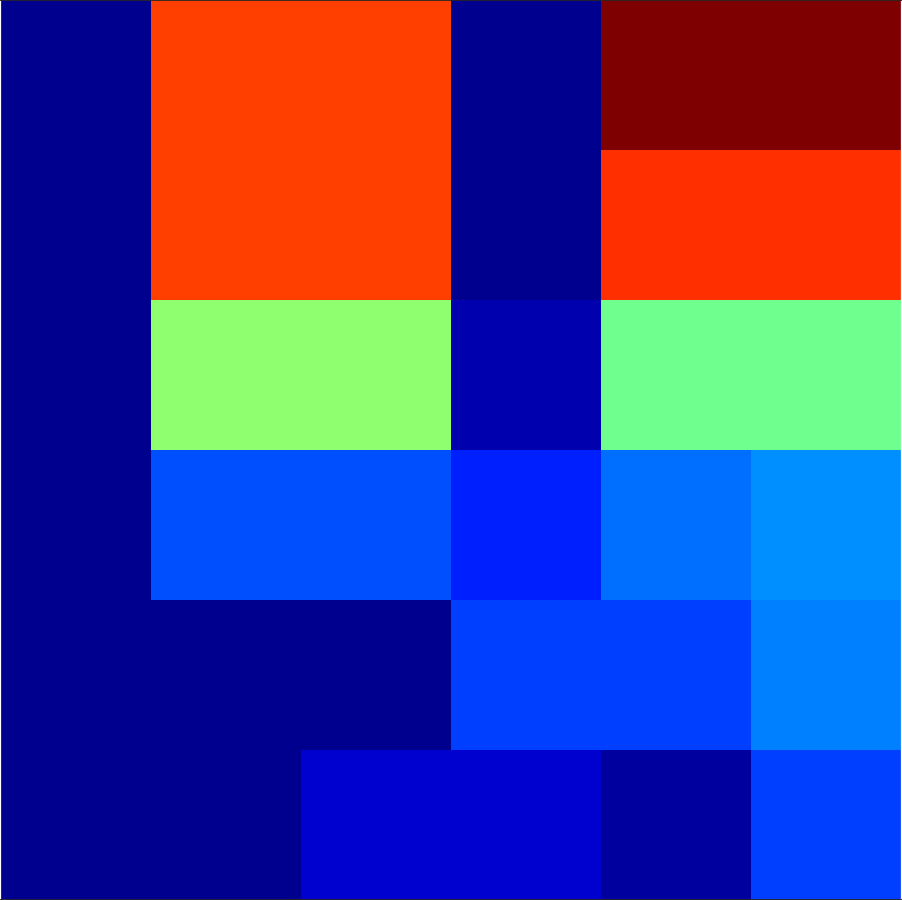}&
\includegraphics[width=0.11\textwidth]{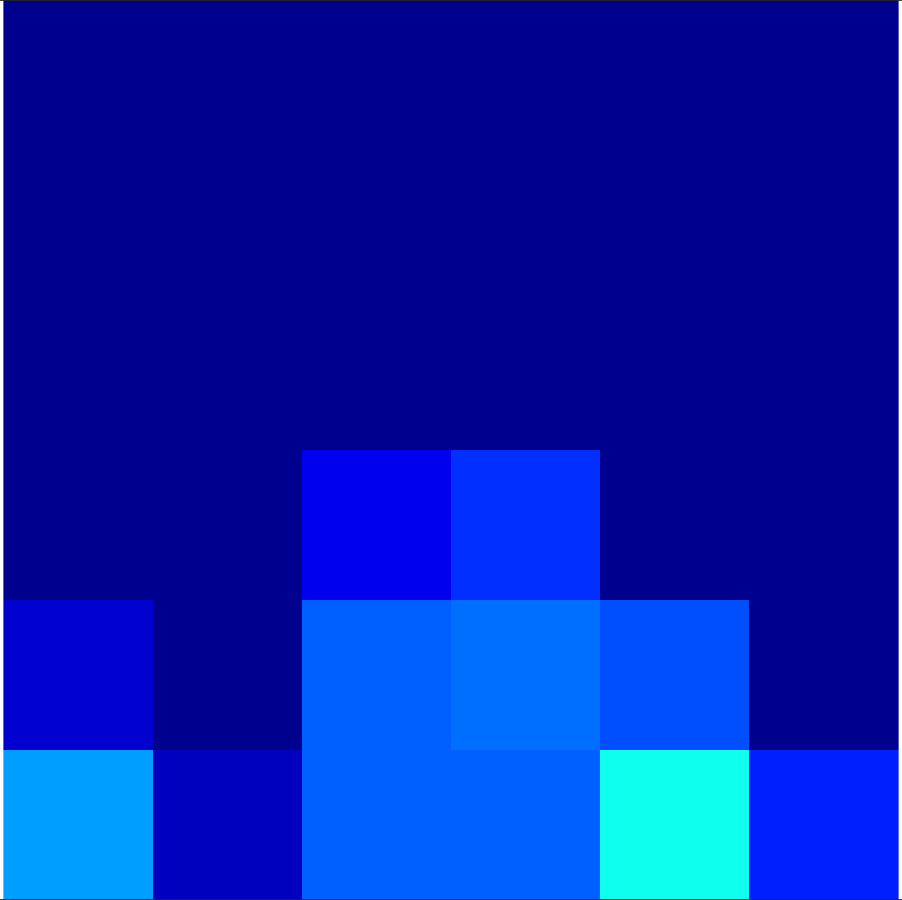}&
\includegraphics[width=0.11\textwidth]{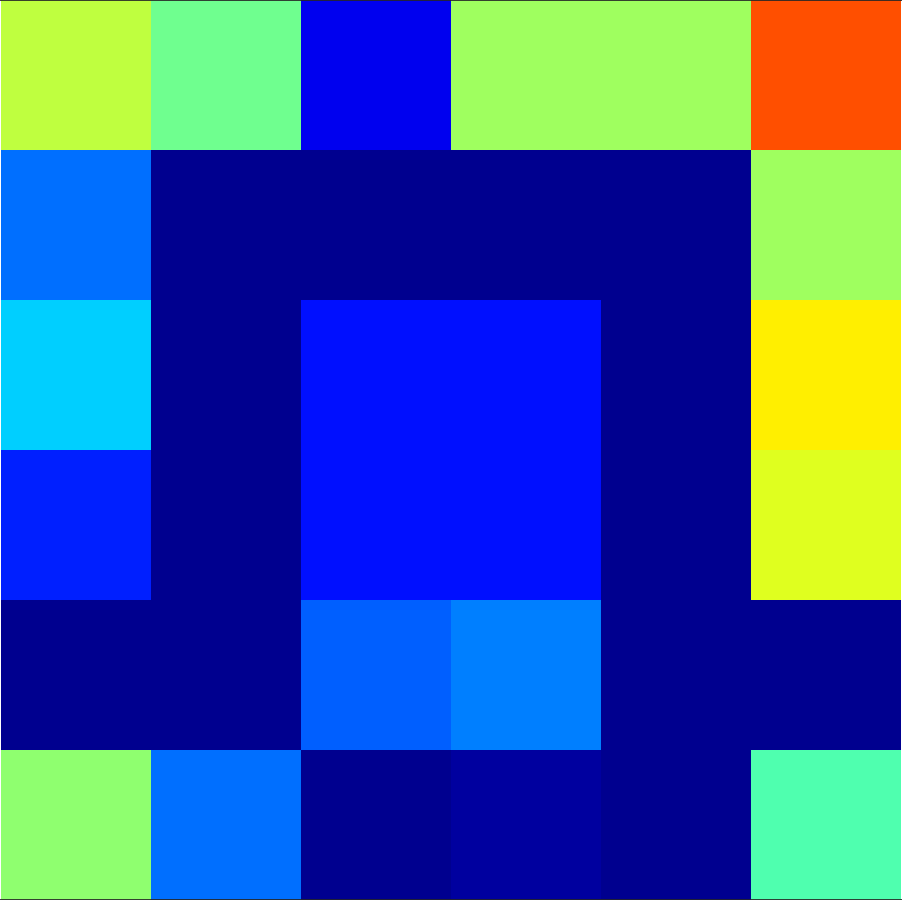}\\
\includegraphics[width=0.11\textwidth , height=0.11\textwidth]{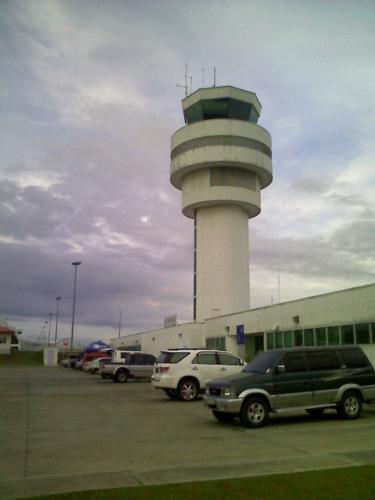}&
\includegraphics[width=0.11\textwidth]{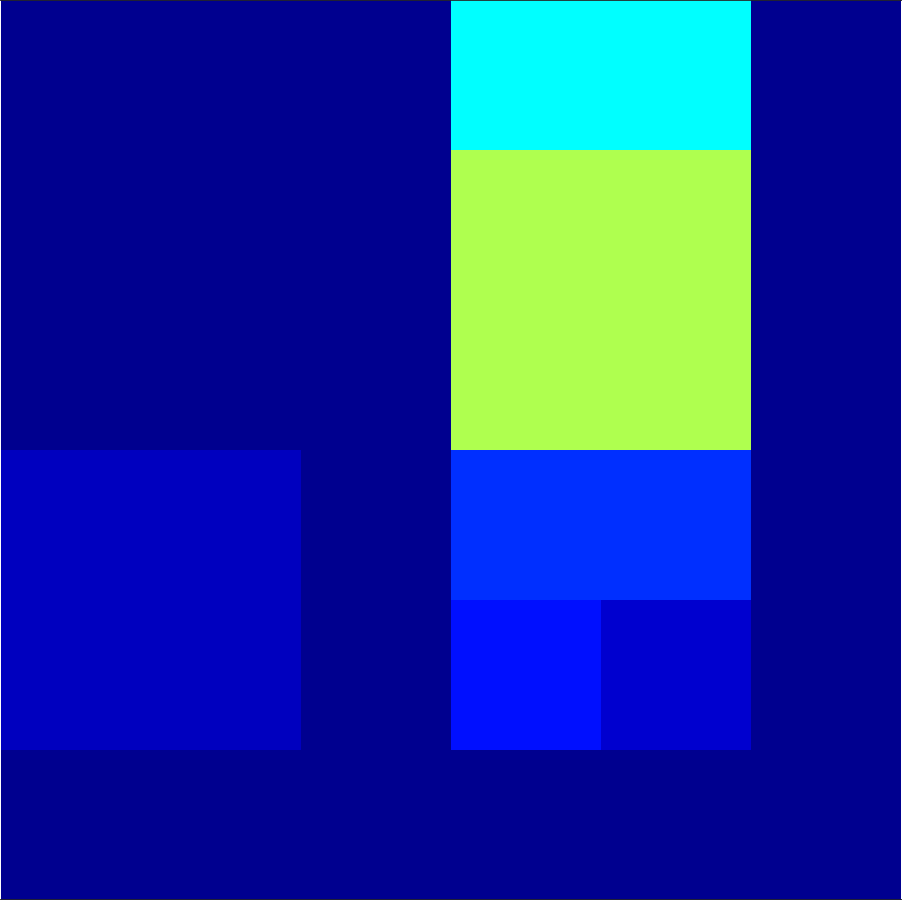}&
\includegraphics[width=0.11\textwidth]{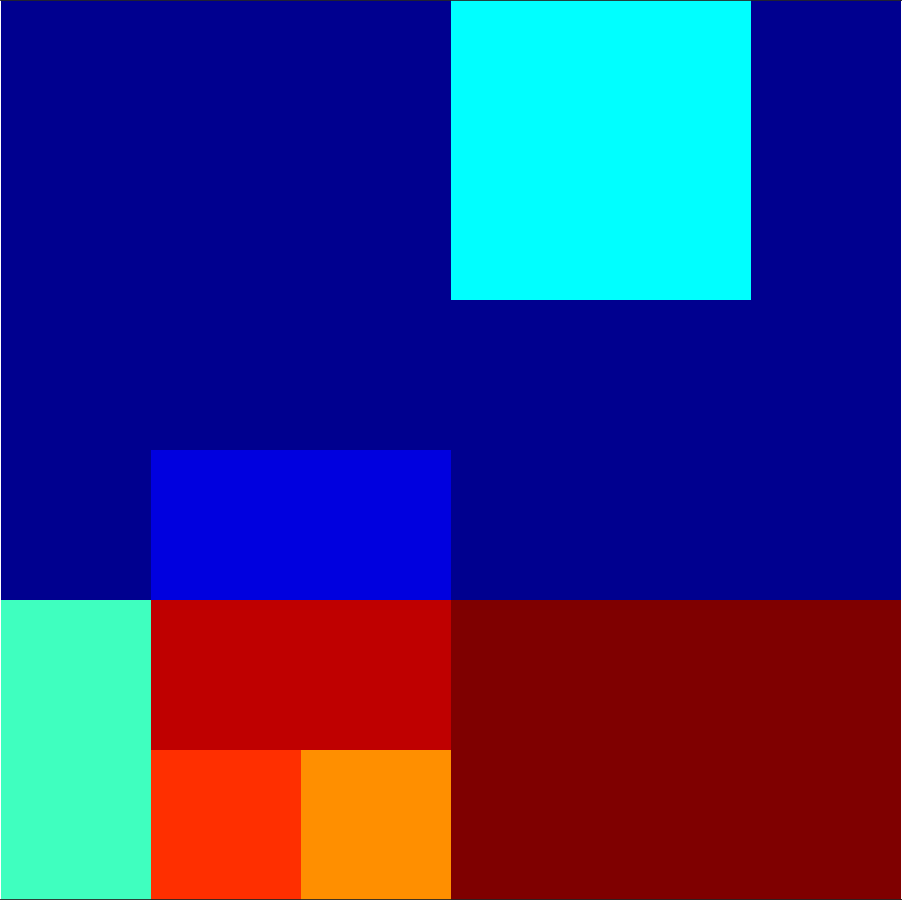}&
\includegraphics[width=0.11\textwidth]{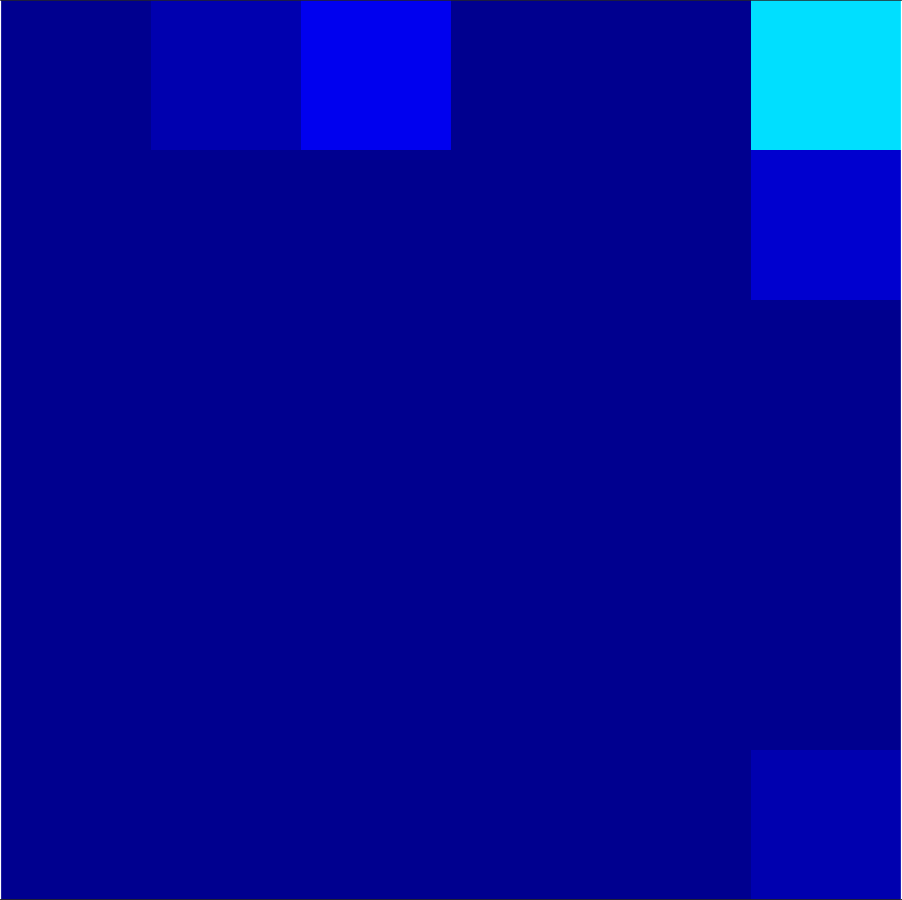}\\
\includegraphics[width=0.11\textwidth , height=0.11\textwidth]{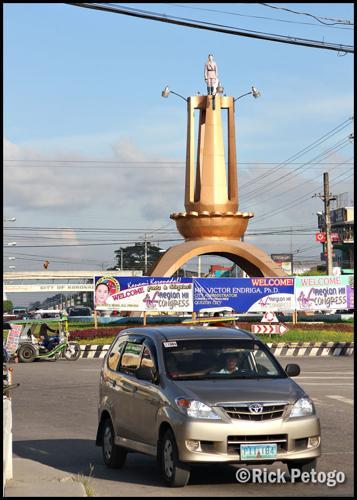}&
\includegraphics[width=0.11\textwidth]{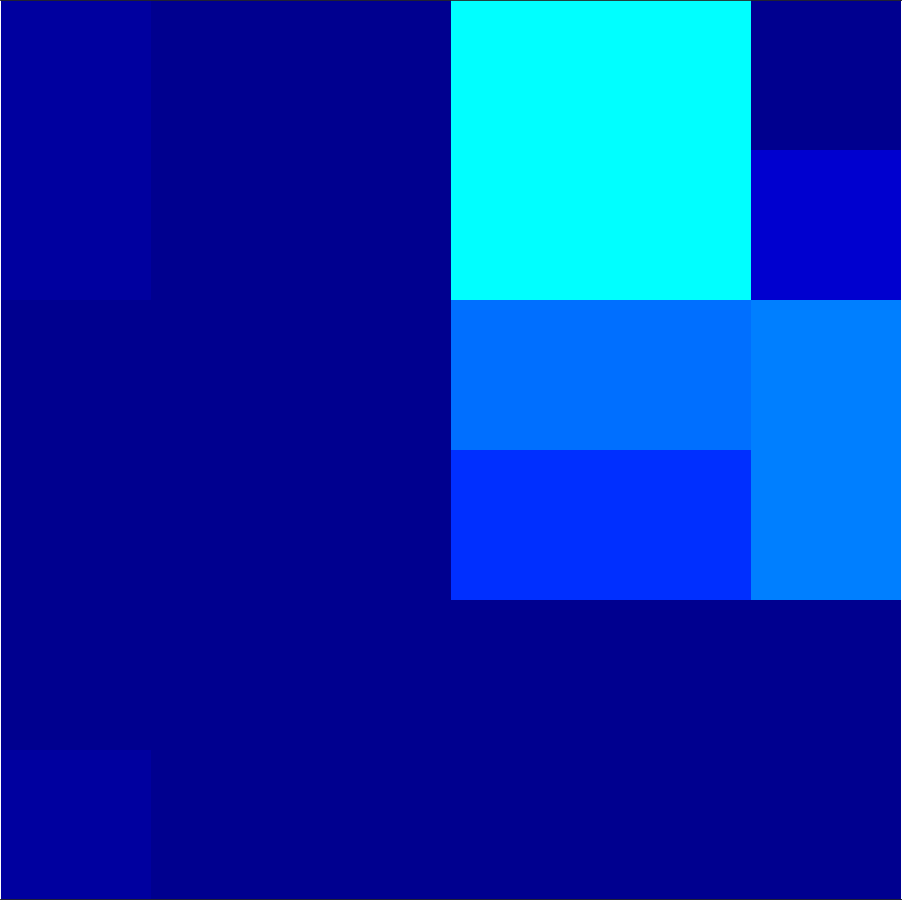}&
\includegraphics[width=0.11\textwidth]{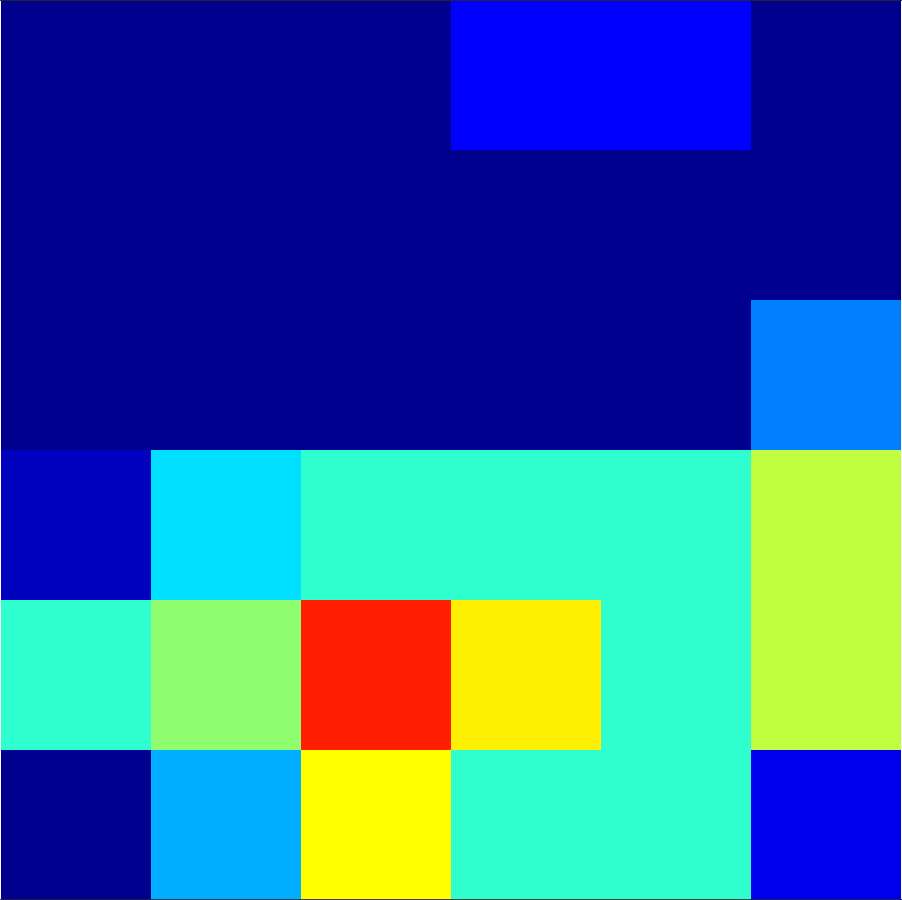}&
\includegraphics[width=0.11\textwidth]{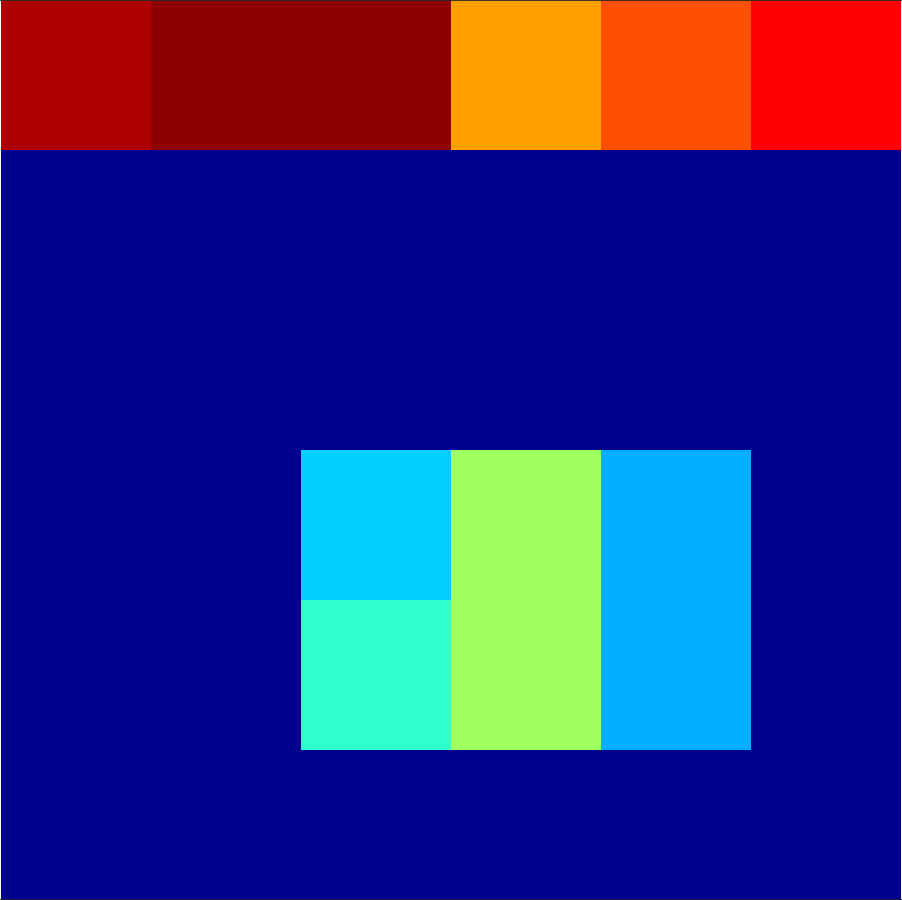}\\
\includegraphics[width=0.11\textwidth , height=0.11\textwidth]{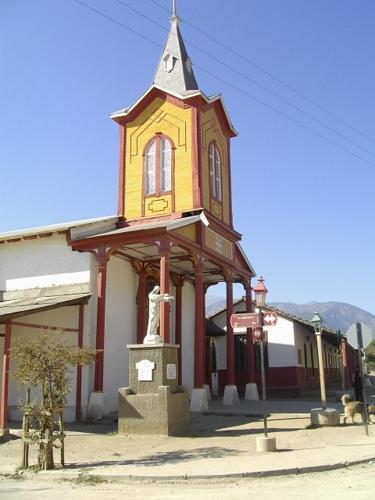}&
\includegraphics[width=0.11\textwidth]{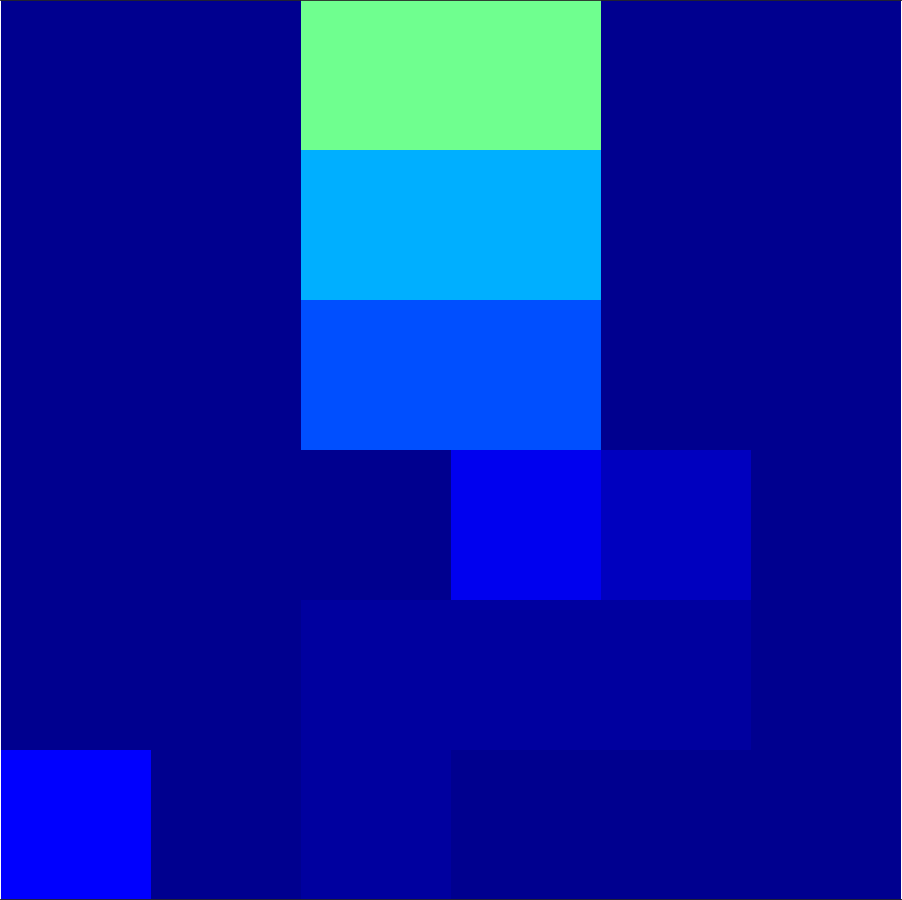}&
\includegraphics[width=0.11\textwidth]{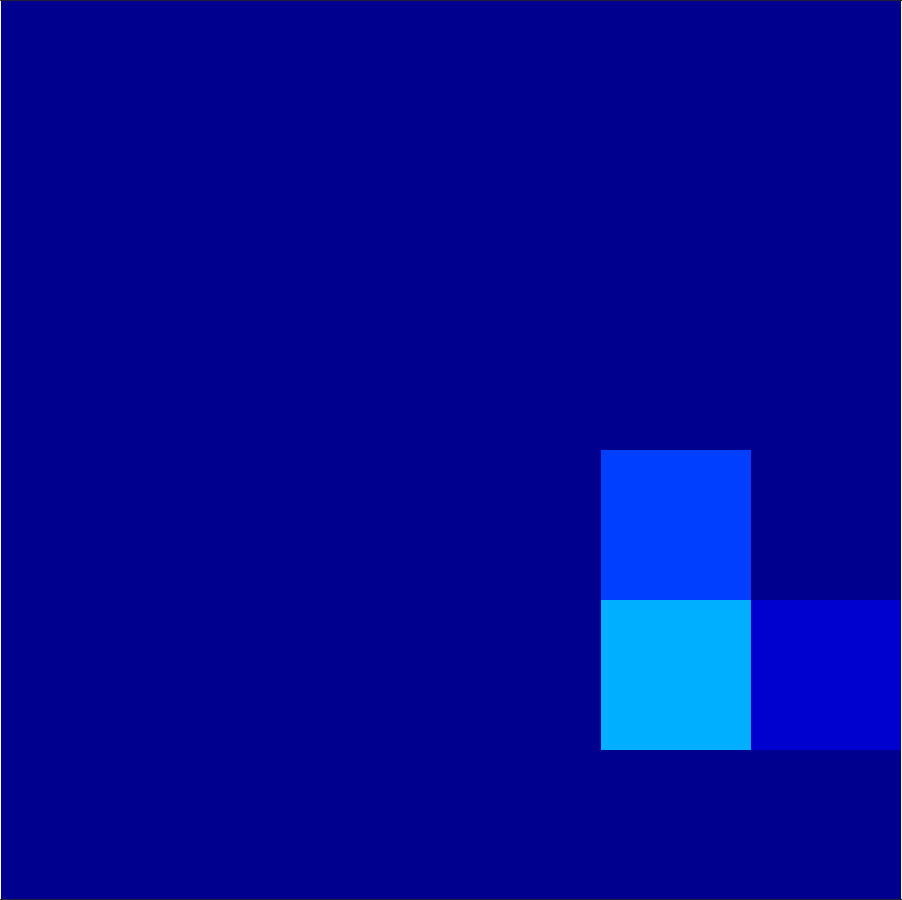}&
\includegraphics[width=0.11\textwidth]{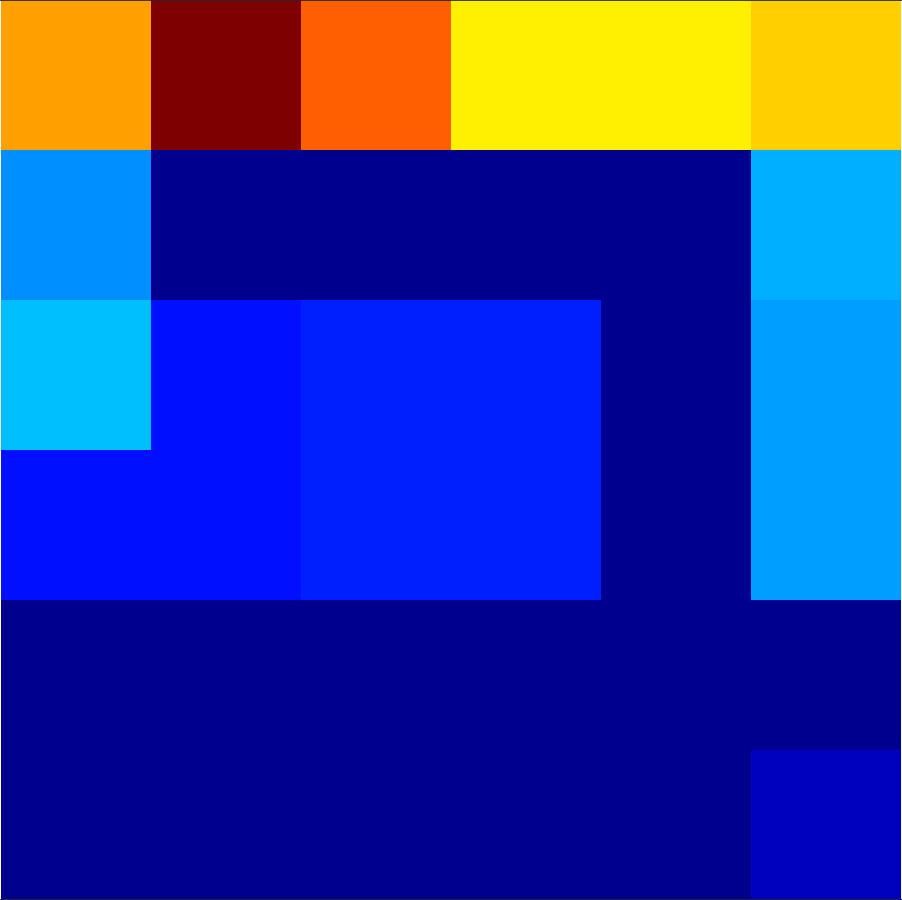}\\
\includegraphics[width=0.11\textwidth , height=0.11\textwidth]{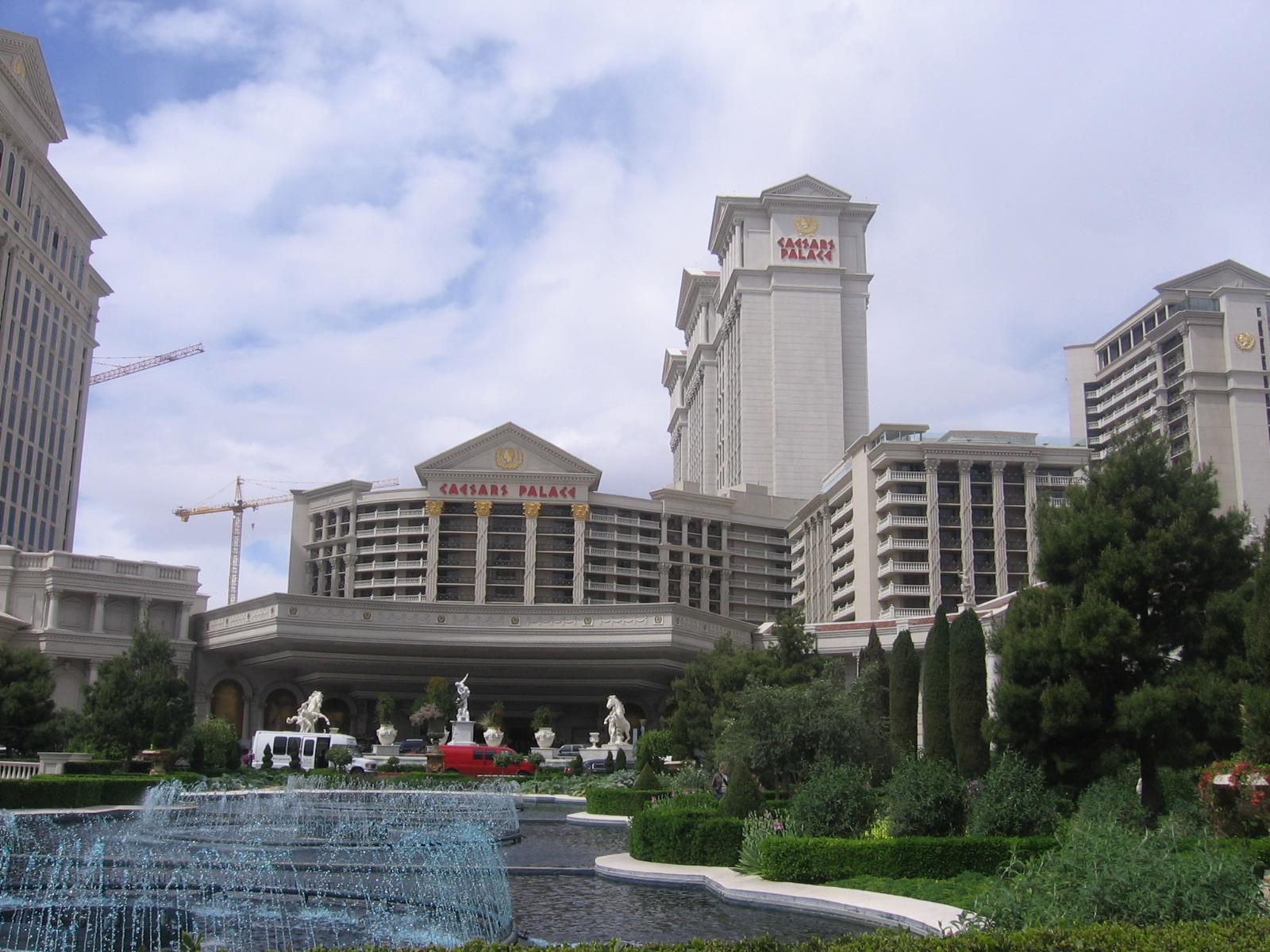}&
\includegraphics[width=0.11\textwidth]{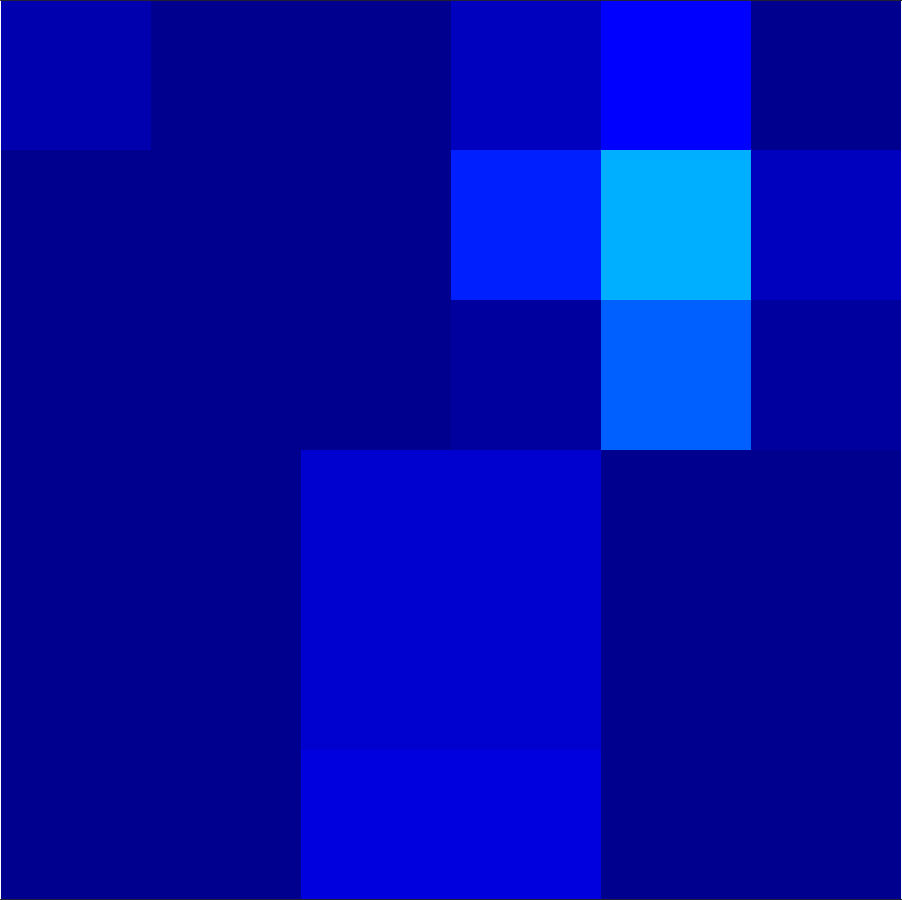}&
\includegraphics[width=0.11\textwidth]{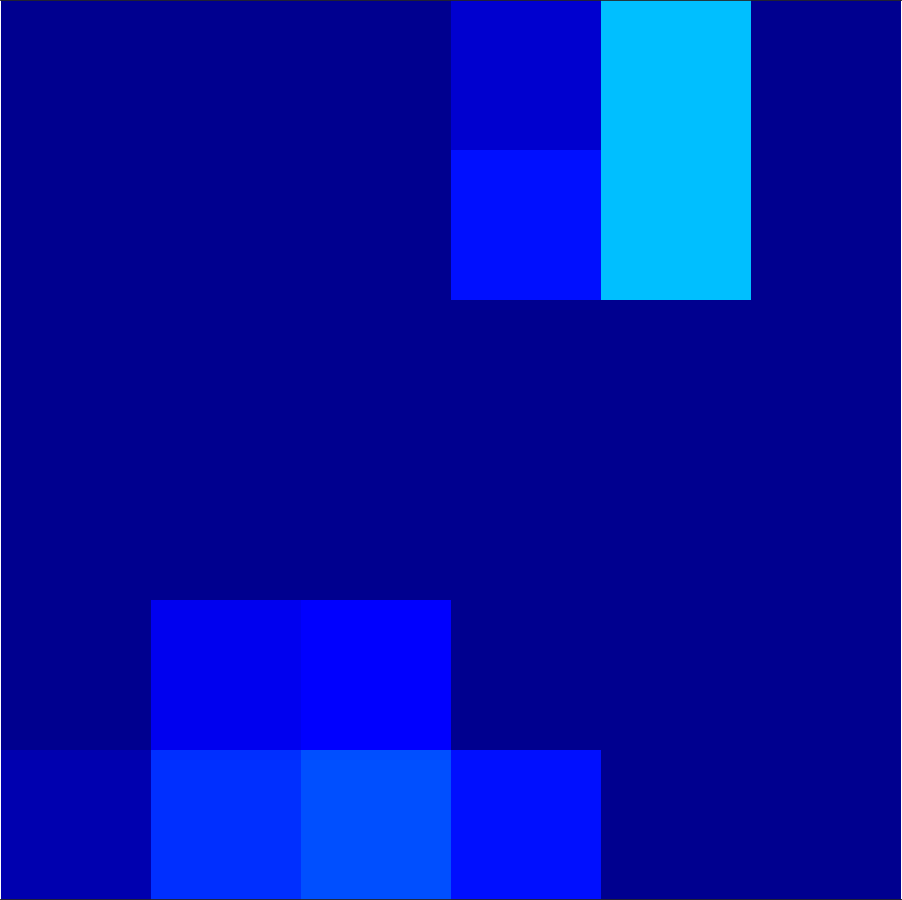}&
\includegraphics[width=0.11\textwidth]{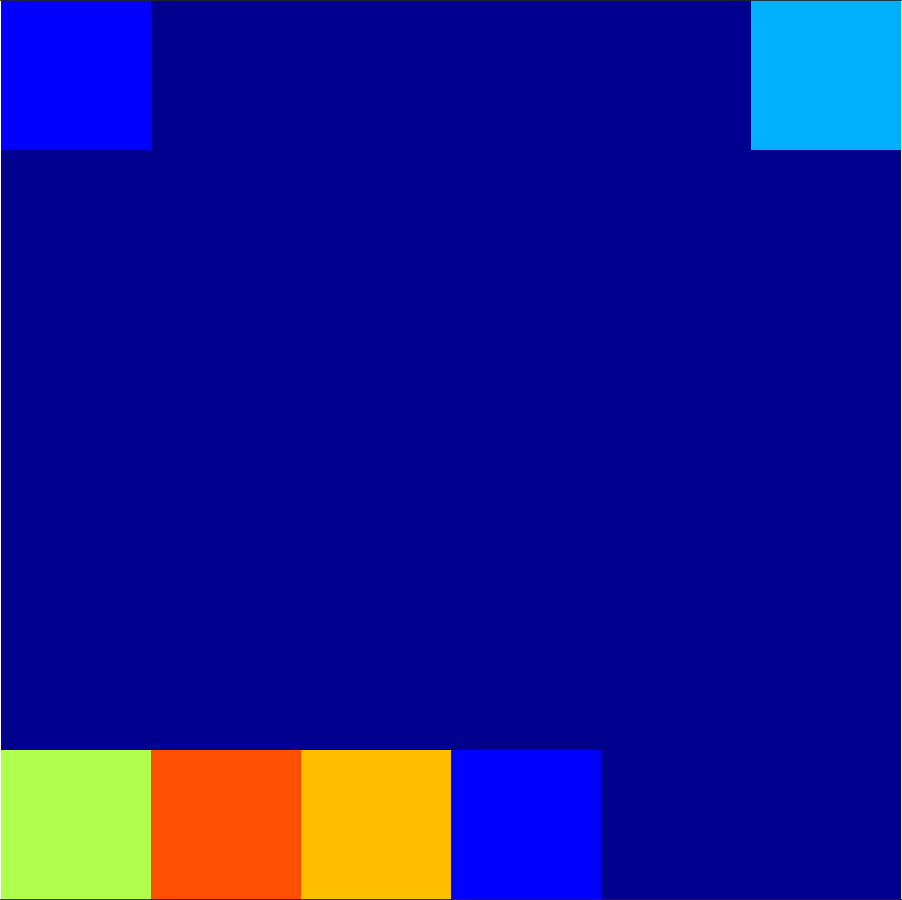}\\
\end{tabular}
\caption{Illustration of semantic information captured by each feature map of pool5 layer using CNN trained on {\em Places} dataset. Each column shows a selected feature map of that layer. All columns are normalized separately and have the same scale. The semantic attributes for each feature map are determined empirically. Note that not only each feature map localizes the concepts, but the magnitude of response is correlated to the scale of each semantic attribute, \ie when tower is seen at smaller scale the number of high activation cells is smaller.}
\label{fig:SemanticFeatureMaps}
\end{figure}
\section{Proposed Method}

Inspired by \cite{Zhou-ICLR2014}, \cite{ZeilerFergus-ECCV2014}, and \cite{DarrellSemSeg2014}, we propose a novel efficient CNN derived image feature which can be used for both image retrieval and scene categorization. Our proposal is motivated by an observation that the feature maps of later convolutional layers of the existing networks already capture fair amount of semantic attributes. As it is shown in Figure~\ref{fig:overview}, each layer consists of $K$ 2D feature maps where each feature map often capturing specific aspect of the image such as the color, object category, or attributes, while preserving the spatial information at coarse resolution. For example, pool5 layer on pre-trained CNNs on {\em ImageNet}~\cite{Krizhevsky-NIPS2012} and {\em Places}~\cite{cnnPlaces-NIPS2014} consists of 256 feature maps where the resolution of each of the feature maps is $13 \times 13$ and $6 \times 6$ respectively. Therefore, the feature maps at this layer preserves spatial information at the resolution of $13 \times 13$ and $6 \times 6$. While earlier layers captures rudimentary concepts such as lines, circles, and stripes, the feature maps in deeper layers can identify more sophisticated concepts. It has been demonstrated that it is possible to identify the meaning of each feature map in a stimuli-based data driven fashion \cite{Zhou-ICLR2014}. Figure~\ref{fig:SemanticFeatureMaps} visualizes some of the feature maps at pool5 layer with their corresponding empirical semantic meaning. As it is shown, feature maps have high responses at the vicinity of the location of that concept. 

\begin{figure}[t]
\begin{tabular}{c@{\hspace{1.5mm}}c@{\hspace{1.5mm}}c@{\hspace{1.5mm}}c}
\includegraphics[width=0.11\textwidth , height=0.11\textwidth]{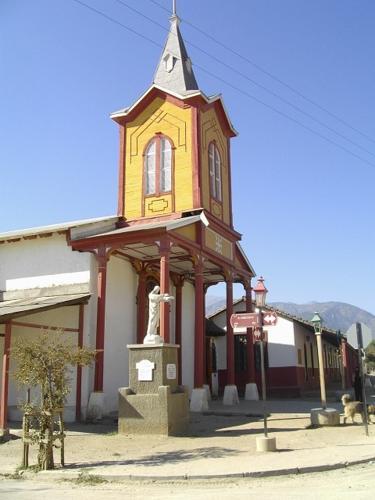}&
\includegraphics[width=0.11\textwidth]{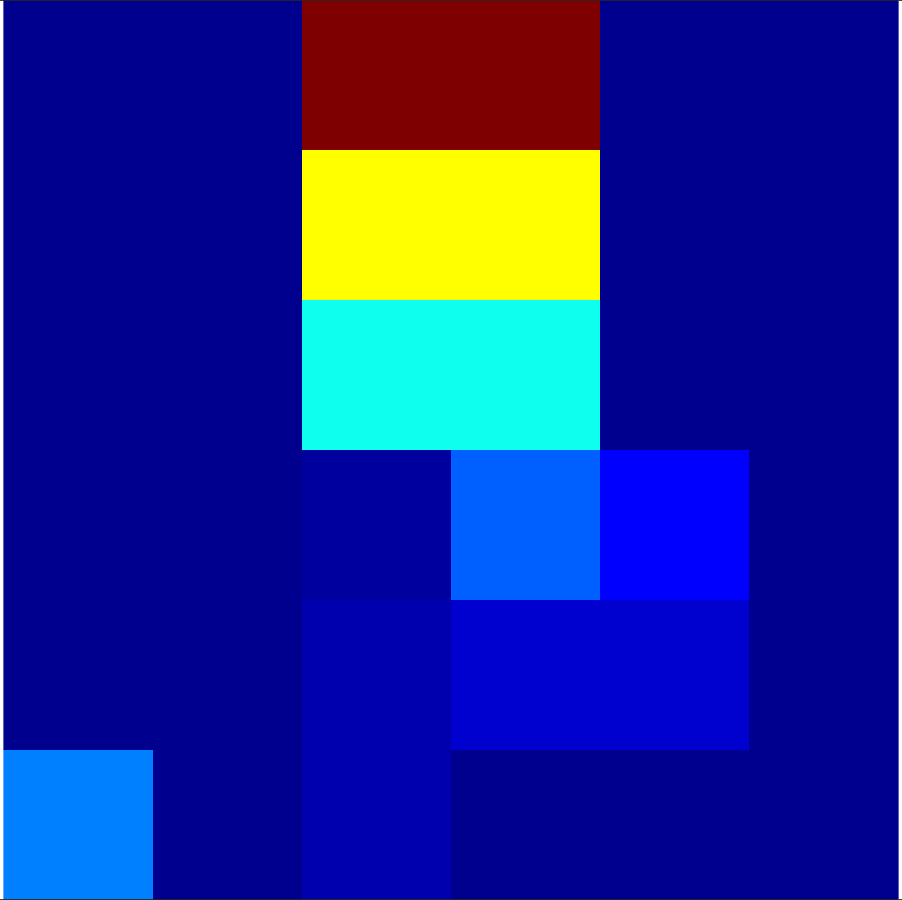}&
\includegraphics[width=0.11\textwidth , height=0.11\textwidth]{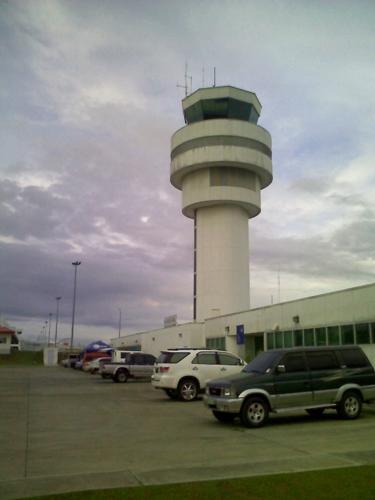}&
\includegraphics[width=0.11\textwidth]{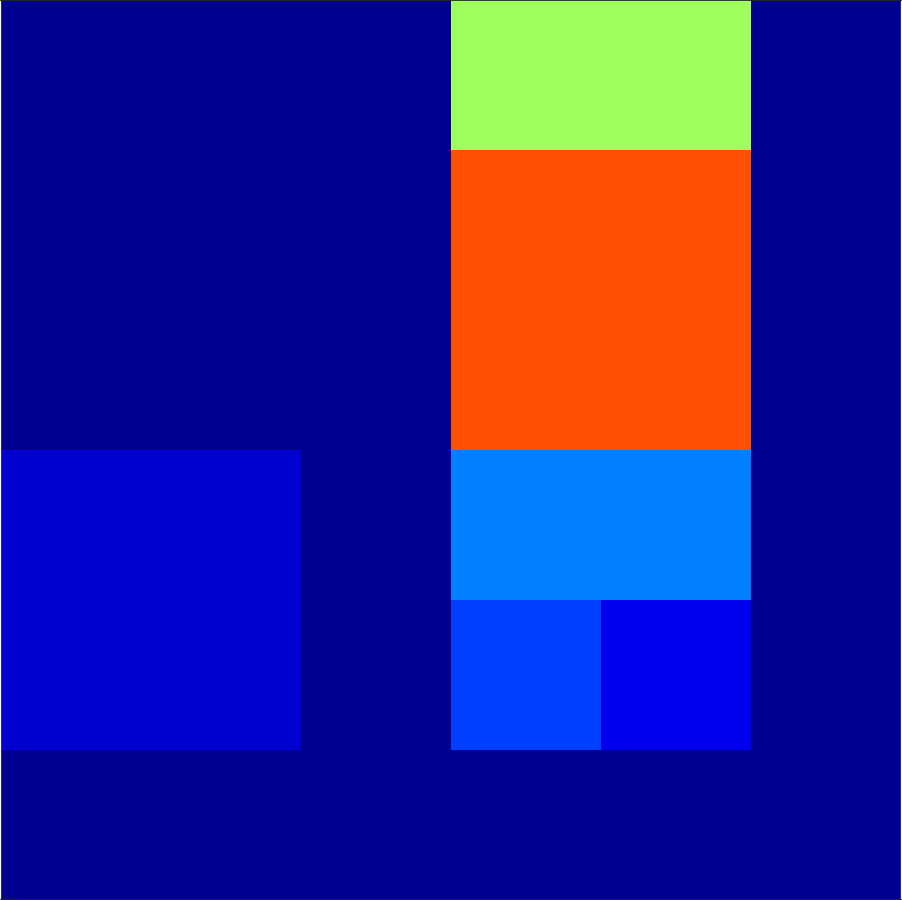}\\
\includegraphics[width=0.11\textwidth , height=0.11\textwidth]{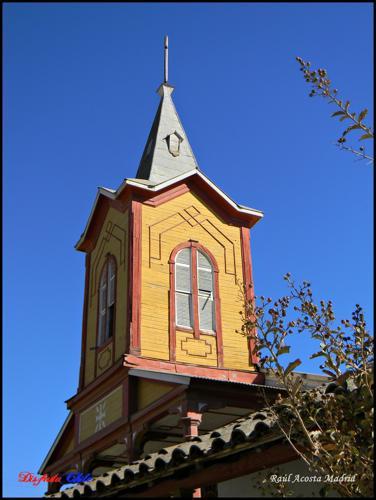}&
\includegraphics[width=0.11\textwidth]{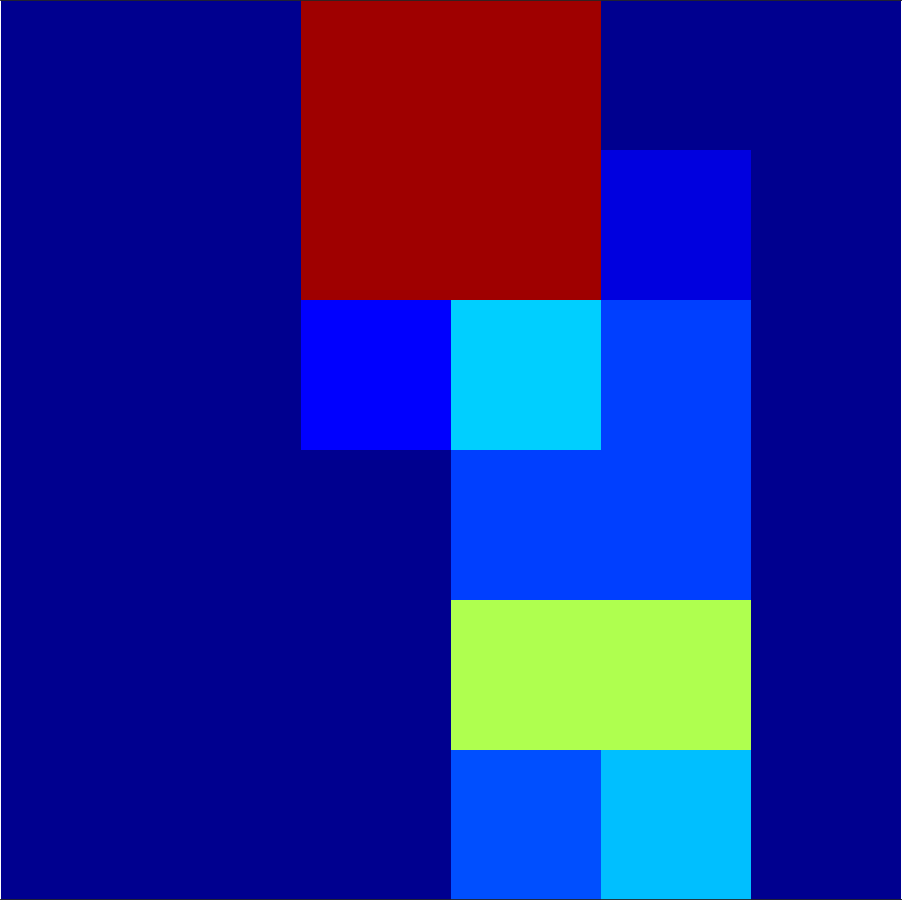}&
\includegraphics[width=0.11\textwidth , height=0.11\textwidth]{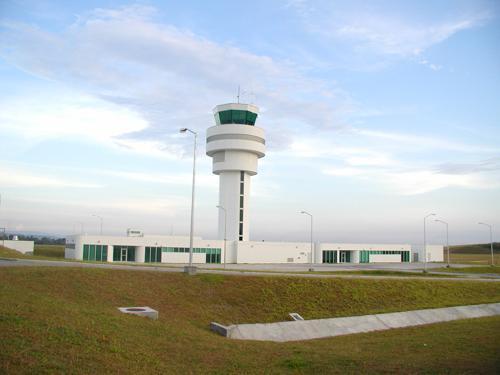}&
\includegraphics[width=0.11\textwidth]{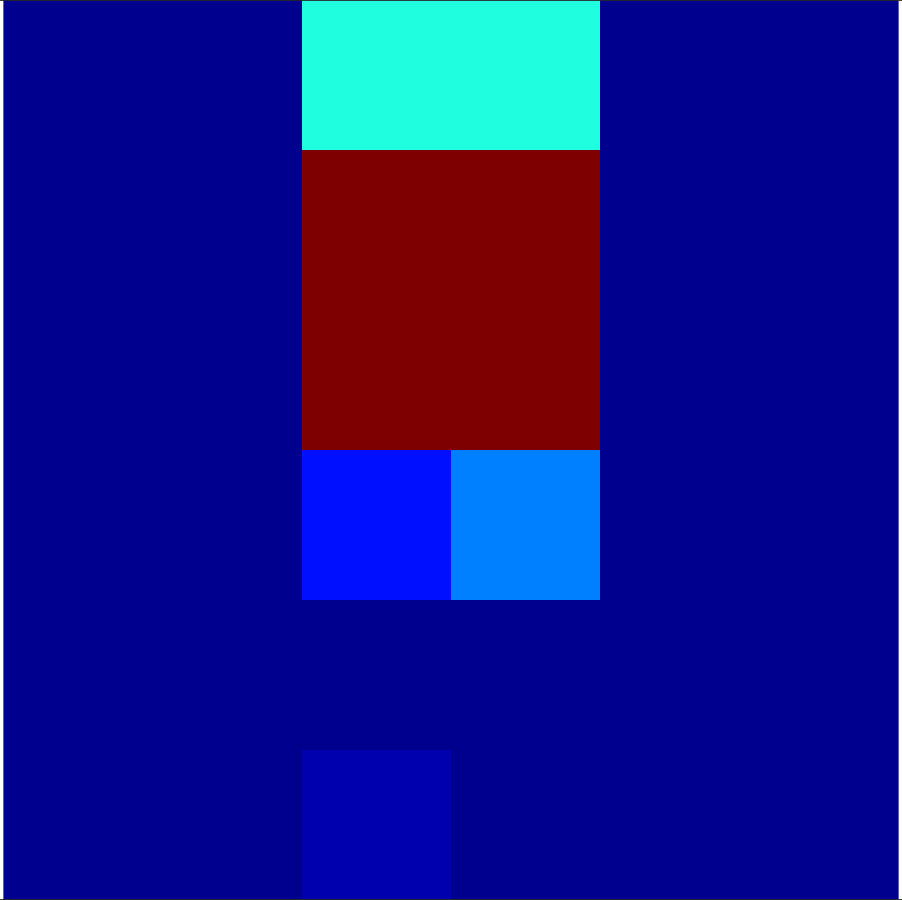}\\
\includegraphics[width=0.11\textwidth , height=0.11\textwidth]{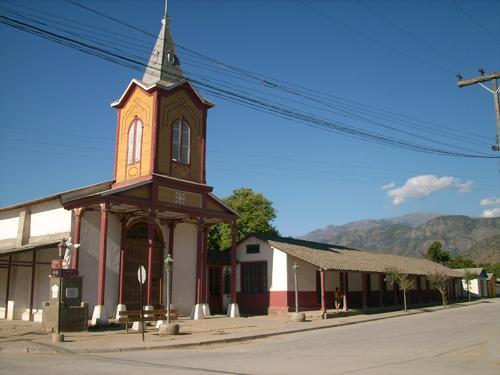}&
\includegraphics[width=0.11\textwidth]{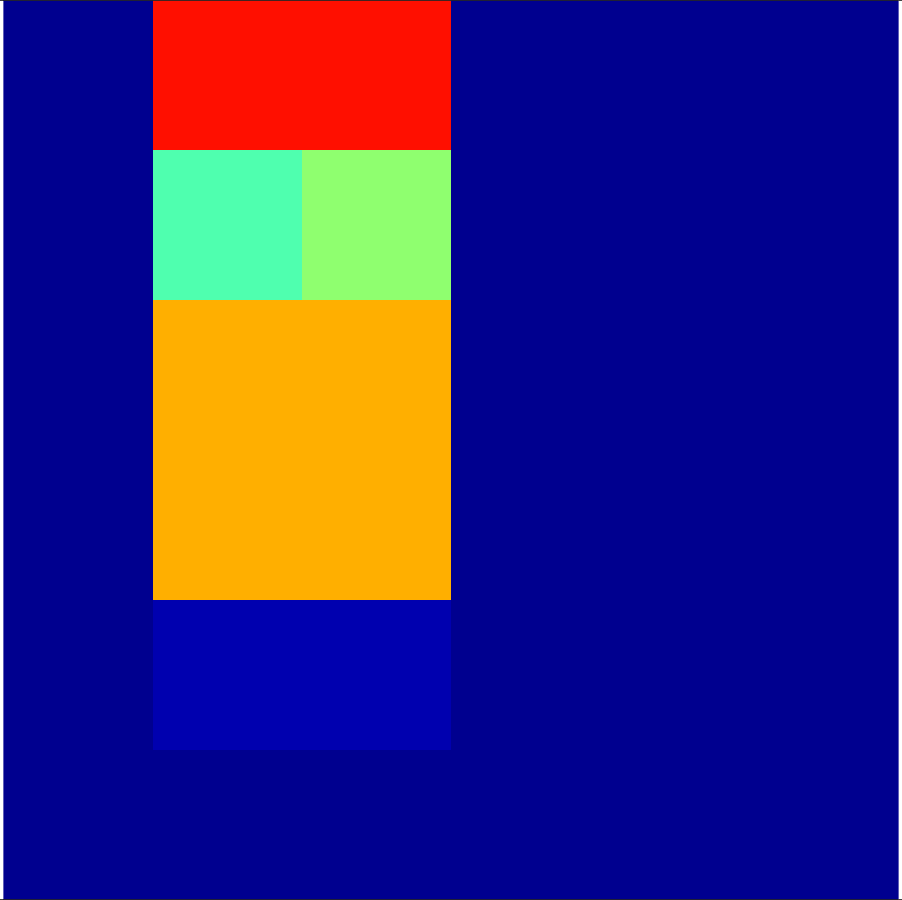}&
\includegraphics[width=0.11\textwidth , height=0.11\textwidth]{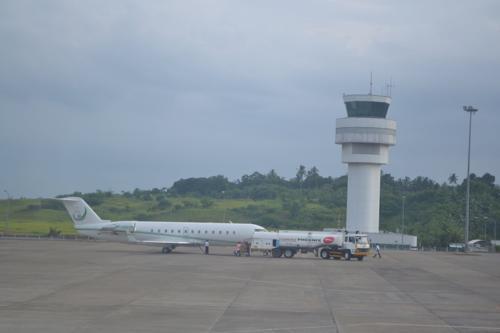}&
\includegraphics[width=0.11\textwidth]{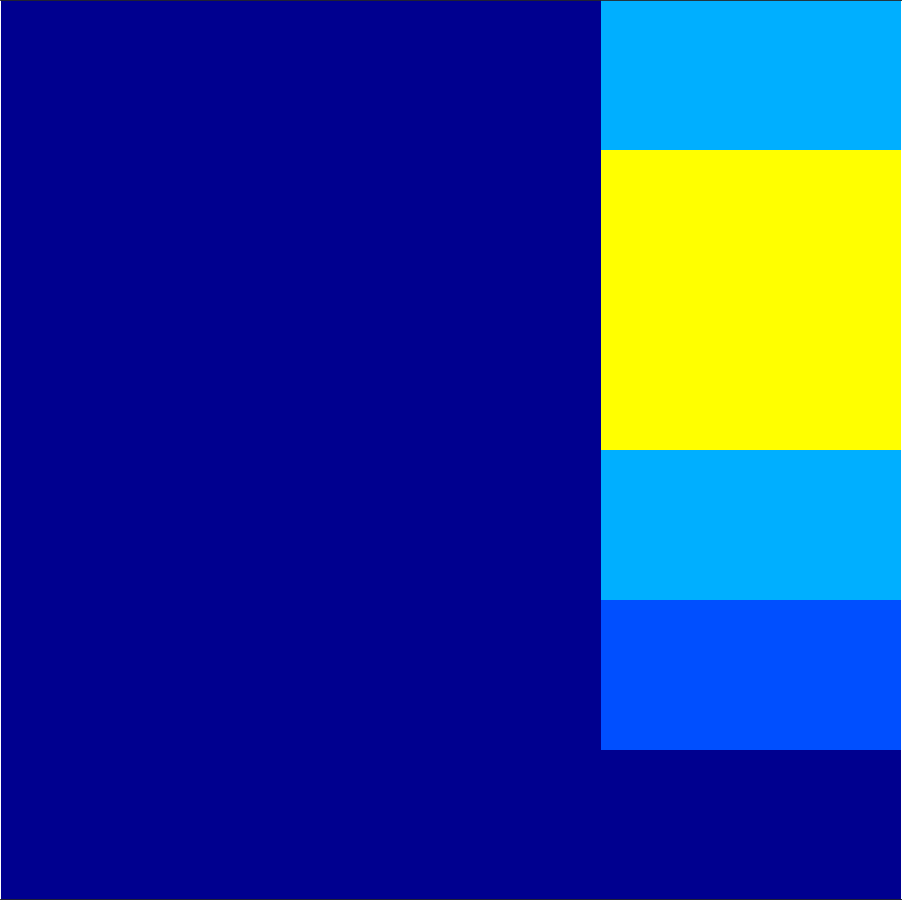}\\
(a) & (b) & (c) & (d)
\end{tabular}
\caption{Effect of translation and scale on pool5 feature maps. (a) and (c) are images of a same place with different translation and scale. (c) and (d) are the feature map for "towerness". Note that the magnitude of feature maps change with the scale change and translation.}
\label{fig:ScaleTranslationEffect}
\end{figure}

We construct the proposed representation by pooling from each feature map of pool5 layer. Therefore, the dimensionality of our representation is linearly proportional to the number of feature maps at pool5 layer, which is 256 in case of {\em ImageNet} and {\em Places} pre-trained convolutional neural networks. The proposed image representation will then be used for retrieval application or classification. We chose to construct the proposed representation from the feature maps in the pool5 because they contain enough information to reconstruct the image by deconvolution \cite{ZeilerFergus-ECCV2014}. Two types of pooling, which are widely used, are max pooling and average pooling \cite{Szegedy2014}. The rationale behind both max and average pooling is to gain invariance to translation over the region where pooling is performed.  However, these two types of pooling do not behave similarly. Max pooling is more invariant to the scale change, since the maximum response of a feature map does not change abruptly with the scale change. Average pooling is more sensitive to the scale change. The downside of max pooling is that in a presence of a distractor in the image which generates high activation in a certain feature map,  (\eg car on the road in Figure~\ref{fig:SemanticFeatureMaps} which is irrelevant to the retrieval of the correct scene),  max pooling is more affected by that activation. In contrast, average pooling is not so sensitive to these type of distractors in the feature maps as it averages the responses over the whole feature map. Figure~\ref{fig:ScaleTranslationEffect} shows the response of most active feature maps at pool5 layer for the images of the same place but with notable translation or scale variations. Note that the maximum of each feature map does not change dramatically with the scale but the averages of the feature maps are related to the scale of the "towerness" concept.
We propose evaluate the features from the pool5 layer of the network followed by following 3 pooling strategies, yielding different image representation: 
\begin{itemize}
\item  Max Pooling yielding 256-dimensional feature where i$^{th}$ element is the result of max pooling on i$^{th}$ feature map at pool5 layer; 
\item Average Pooling yielding with 256-dimensional feature such that i$^{th}$  element is the result of average pooling on the i$^{th}$  feature map at pool5 layer;
\item Hybrid Pooling yielding  512-dimensional feature where the representation is the concatenation of max pooling and average pooling representation.
\end{itemize} 

We also perform whitening of each dimension of the final representation separately such that all the dimensions of the representation have zero mean and unit variance to prevent some feature maps with large responses having a large effect on the final representation. Our method is considerably more efficient than \cite{Lazebnik-arXiv2014} where the authors compute  fc7 features on the image itself, 25 patches of $128\times128$ pixels, and 49 patches of $64\times64$ pixels, which results in running the convolutional network for each image 75 times. Since combing all 3 scale levels yield 12,288 dimensional features vector, authors further experiment with PCA dimensionality reduction, pooling and quantizations to reduce the dimensionality of the resulting features. These additional techniques affect favorably image retrieval problem, but for classification the high-dimensional features perform best.  
Our representation is substantially simpler, low-dimensional and is computed by passing each image through the convolutional neural network once.

\begin{figure*}
\centering
\begin{tabular}{@{\hspace{0mm}}c@{\hspace{-1.5mm}}c@{\hspace{1.5mm}}c@{\hspace{1.5mm}}c}
\begin{tabular}{c}
\includegraphics[width=0.09\textwidth, height=0.09\textwidth]{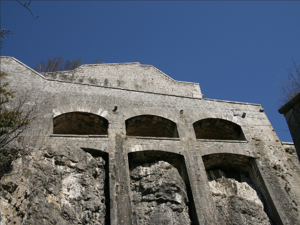} \\
\includegraphics[width=0.09\textwidth, height=0.09\textwidth]{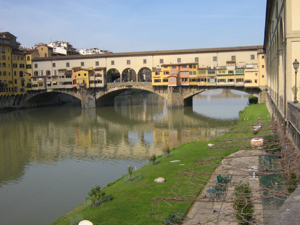} \\
\includegraphics[width=0.09\textwidth, height=0.09\textwidth]{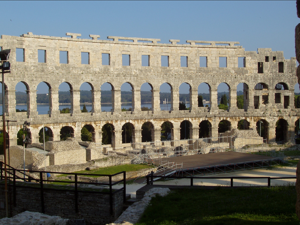} \\
\includegraphics[width=0.09\textwidth, height=0.09\textwidth]{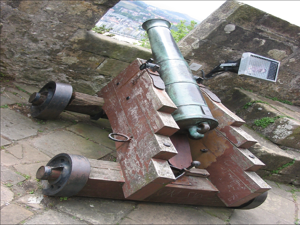} \\
\end{tabular}& 
\begin{tabular}{|@{\hspace{1.5mm}}c@{\hspace{1.5mm}}c@{\hspace{1.5mm}}c@{\hspace{1.5mm}}|}
\hline
\includegraphics[width=0.09\textwidth, height=0.09\textwidth]{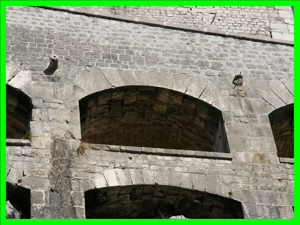} & 
\includegraphics[width=0.09\textwidth, height=0.09\textwidth]{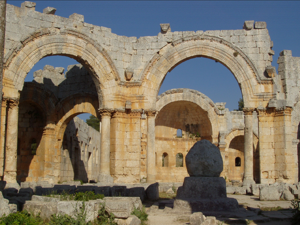} &
\includegraphics[width=0.09\textwidth, height=0.09\textwidth]{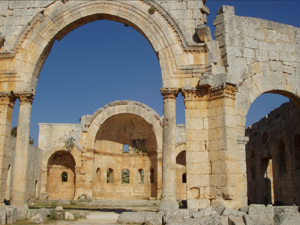} \\

\includegraphics[width=0.09\textwidth, height=0.09\textwidth]{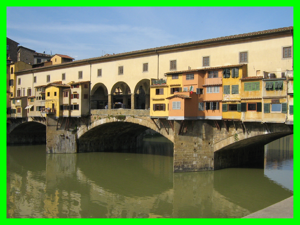} & 
\includegraphics[width=0.09\textwidth, height=0.09\textwidth]{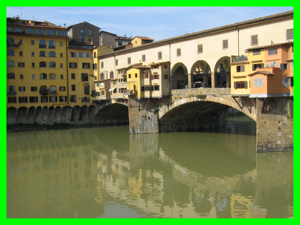} &
\includegraphics[width=0.09\textwidth, height=0.09\textwidth]{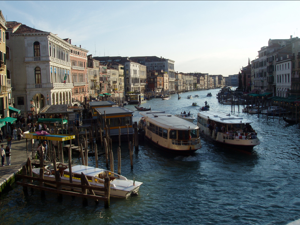} \\

\includegraphics[width=0.09\textwidth, height=0.09\textwidth]{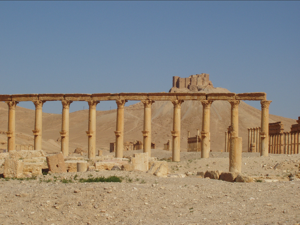} & 
\includegraphics[width=0.09\textwidth, height=0.09\textwidth]{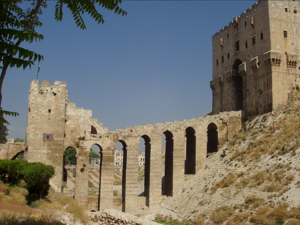} &
\includegraphics[width=0.09\textwidth, height=0.09\textwidth]{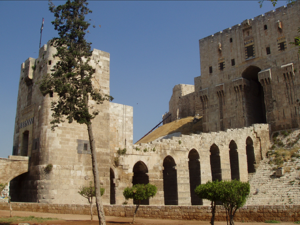} \\

\includegraphics[width=0.09\textwidth, height=0.09\textwidth]{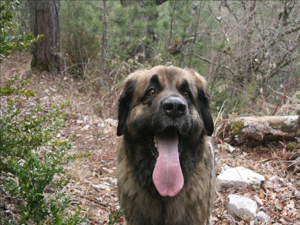} & 
\includegraphics[width=0.09\textwidth, height=0.09\textwidth]{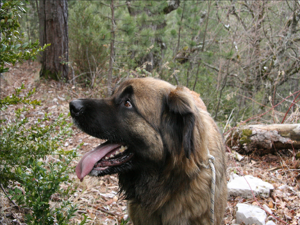} &
\includegraphics[width=0.09\textwidth, height=0.09\textwidth]{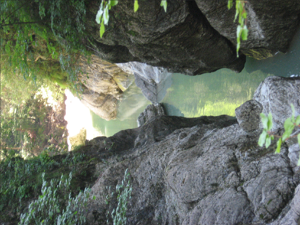} \\
\hline
\end{tabular} &
\begin{tabular}{|@{\hspace{1.5mm}}c@{\hspace{1.5mm}}c@{\hspace{1.5mm}}c@{\hspace{1.5mm}}|}
\hline
\includegraphics[width=0.09\textwidth, height=0.09\textwidth]{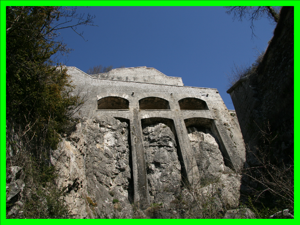} & 
\includegraphics[width=0.09\textwidth, height=0.09\textwidth]{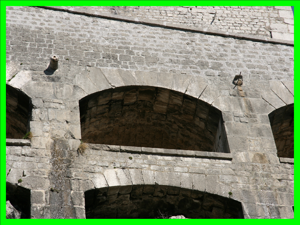} &
\includegraphics[width=0.09\textwidth, height=0.09\textwidth]{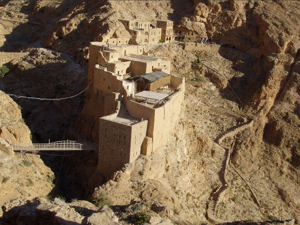} \\

\includegraphics[width=0.09\textwidth, height=0.09\textwidth]{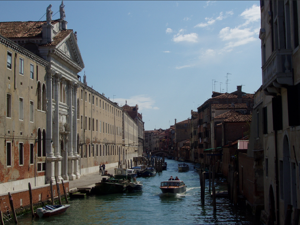} & 
\includegraphics[width=0.09\textwidth, height=0.09\textwidth]{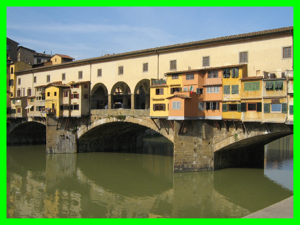} &
\includegraphics[width=0.09\textwidth, height=0.09\textwidth]{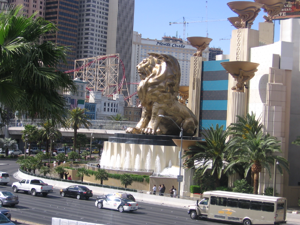} \\

\includegraphics[width=0.09\textwidth, height=0.09\textwidth]{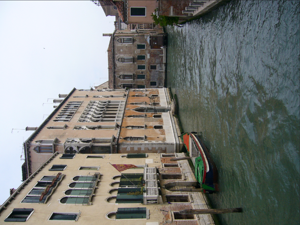} & 
\includegraphics[width=0.09\textwidth, height=0.09\textwidth]{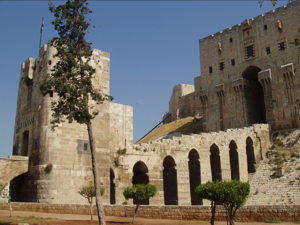} &
\includegraphics[width=0.09\textwidth, height=0.09\textwidth]{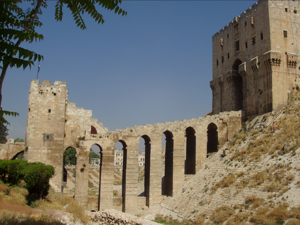} \\

\includegraphics[width=0.09\textwidth, height=0.09\textwidth]{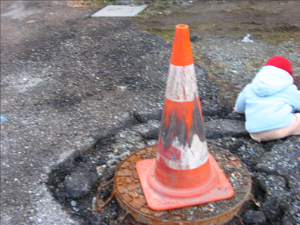} & 
\includegraphics[width=0.09\textwidth, height=0.09\textwidth]{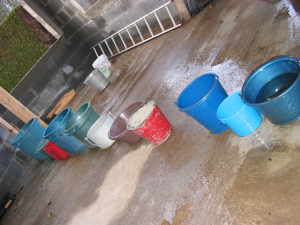} &
\includegraphics[width=0.09\textwidth, height=0.09\textwidth]{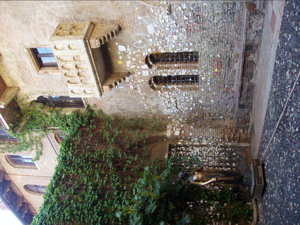} \\
\hline
\end{tabular} &
\begin{tabular}{|@{\hspace{1.5mm}}c@{\hspace{1.5mm}}c@{\hspace{1.5mm}}c@{\hspace{1.5mm}}|}
\hline
\includegraphics[width=0.09\textwidth, height=0.09\textwidth]{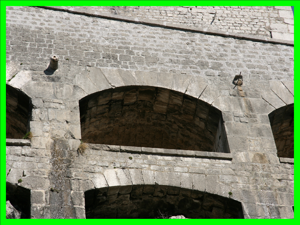} & 
\includegraphics[width=0.09\textwidth, height=0.09\textwidth]{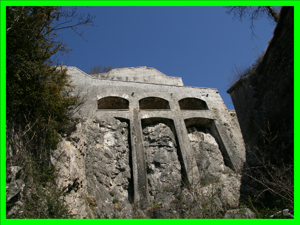} &
\includegraphics[width=0.09\textwidth, height=0.09\textwidth]{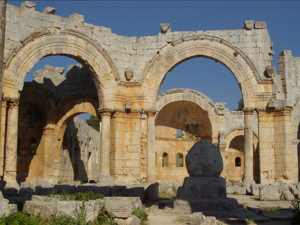} \\

\includegraphics[width=0.09\textwidth, height=0.09\textwidth]{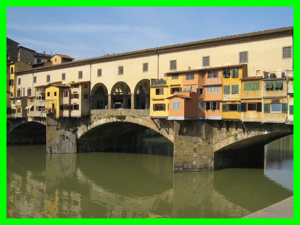} & 
\includegraphics[width=0.09\textwidth, height=0.09\textwidth]{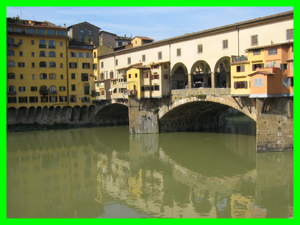} &
\includegraphics[width=0.09\textwidth, height=0.09\textwidth]{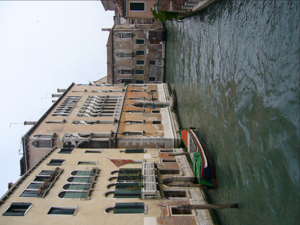} \\

\includegraphics[width=0.09\textwidth, height=0.09\textwidth]{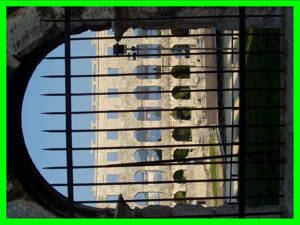} & 
\includegraphics[width=0.09\textwidth, height=0.09\textwidth]{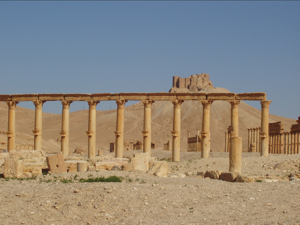} &
\includegraphics[width=0.09\textwidth, height=0.09\textwidth]{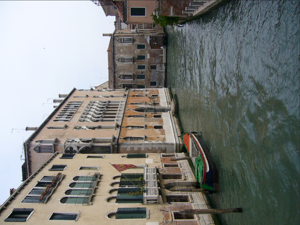} \\

\includegraphics[width=0.09\textwidth, height=0.09\textwidth]{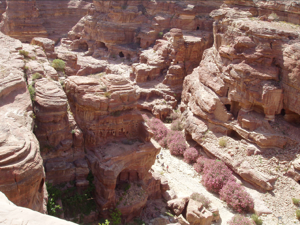} & 
\includegraphics[width=0.09\textwidth, height=0.09\textwidth]{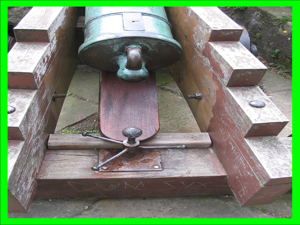} &
\includegraphics[width=0.09\textwidth, height=0.09\textwidth]{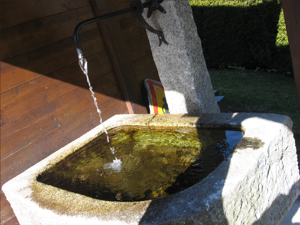} \\
\hline
\end{tabular} \\
Query Image & Average Pooling & Max Pooling & Hybrid Pooling
\end{tabular}
\caption{Qualitative Comparison of Average Pooling, Max Pooling, and Hybrid pooling on INRIA Holidays dataset. For each query image, top 3 images retrieved by max/average/hybrid pooling are shown from left to right. Correctly retrieved images are surrounded by green rectangle (Best viewed in electronic version). Max pooling is more robust against scale change while average pooling is retrieving images with similar scale. Last two rows are query images where only hybrid pooling is able to retrieve correct images in the top 3 images. The feature representations were whitened but no PCA dimensionality reduction is applied.}
\label{fig:AvgMaxHybridPooling}
\end{figure*}

\begin{figure*}
\centering
\begin{tabular}{@{\hspace{1.5mm}}c@{\hspace{1.5mm}}c@{\hspace{1.5mm}}c@{\hspace{1.5mm}}}
\begin{tabular}{c}
\includegraphics[width=0.1\textwidth, height=0.1\textwidth]{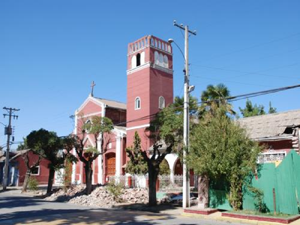} \\
\includegraphics[width=0.1\textwidth, height=0.1\textwidth]{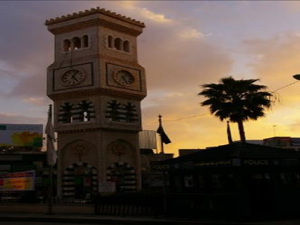} \\
\includegraphics[width=0.1\textwidth, height=0.1\textwidth]{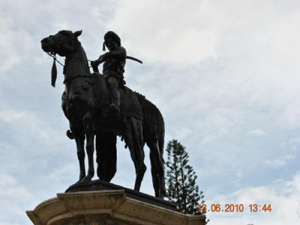} \\
\includegraphics[width=0.1\textwidth, height=0.1\textwidth]{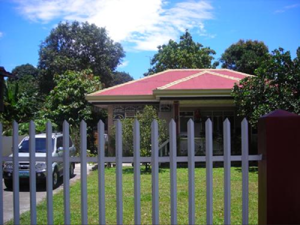} \\
\includegraphics[width=0.1\textwidth, height=0.1\textwidth]{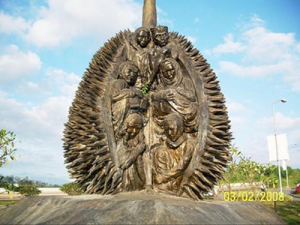} \\
\includegraphics[width=0.1\textwidth, height=0.1\textwidth]{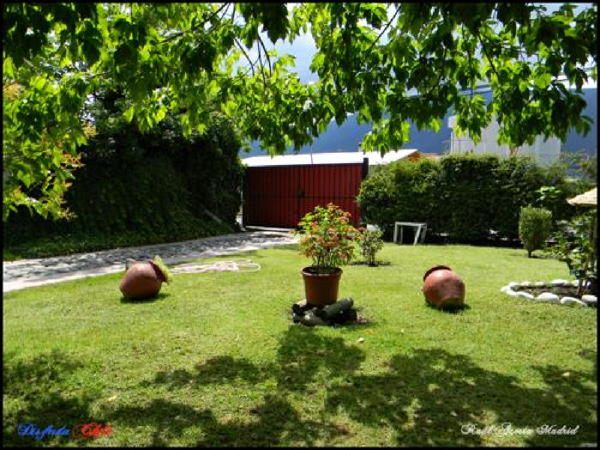} \\
\end{tabular}& 
\begin{tabular}{|@{\hspace{1.5mm}}c@{\hspace{1.5mm}}c@{\hspace{1.5mm}}c@{\hspace{1.5mm}}c@{\hspace{1.5mm}}|}
\hline
\includegraphics[width=0.1\textwidth, height=0.1\textwidth]{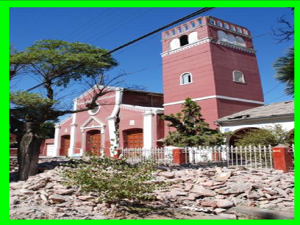} & 
\includegraphics[width=0.1\textwidth, height=0.1\textwidth]{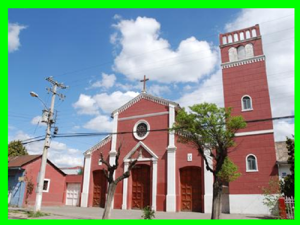} &
\includegraphics[width=0.1\textwidth, height=0.1\textwidth]{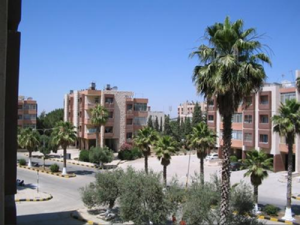} &
\includegraphics[width=0.1\textwidth, height=0.1\textwidth]{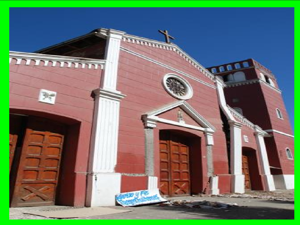}\\
\includegraphics[width=0.1\textwidth, height=0.1\textwidth]{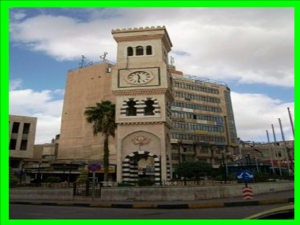} & 
\includegraphics[width=0.1\textwidth, height=0.1\textwidth]{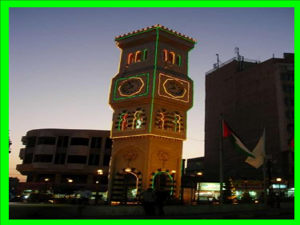} &
\includegraphics[width=0.1\textwidth, height=0.1\textwidth]{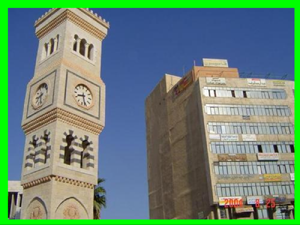} &
\includegraphics[width=0.1\textwidth, height=0.1\textwidth]{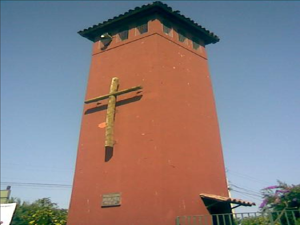}\\
\includegraphics[width=0.1\textwidth, height=0.1\textwidth]{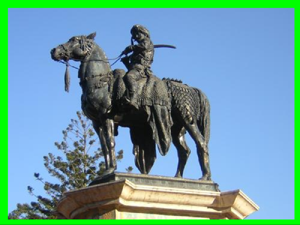} & 
\includegraphics[width=0.1\textwidth, height=0.1\textwidth]{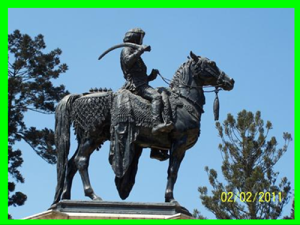} &
\includegraphics[width=0.1\textwidth, height=0.1\textwidth]{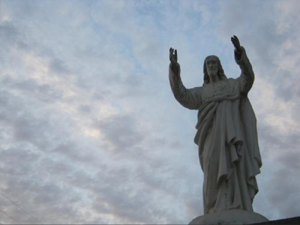} &
\includegraphics[width=0.1\textwidth, height=0.1\textwidth]{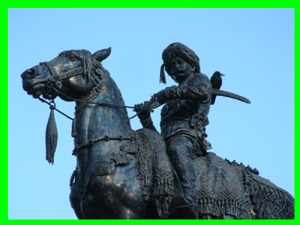}\\
\includegraphics[width=0.1\textwidth, height=0.1\textwidth]{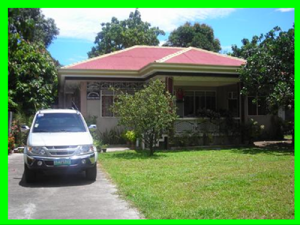} & 
\includegraphics[width=0.1\textwidth, height=0.1\textwidth]{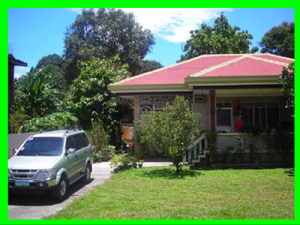} &
\includegraphics[width=0.1\textwidth, height=0.1\textwidth]{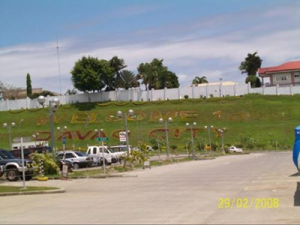} &
\includegraphics[width=0.1\textwidth, height=0.1\textwidth]{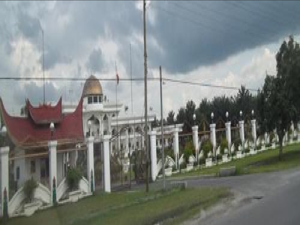}\\
\includegraphics[width=0.1\textwidth, height=0.1\textwidth]{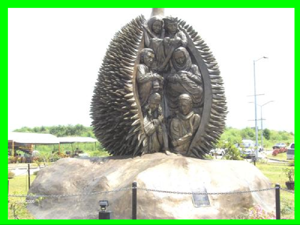} & 
\includegraphics[width=0.1\textwidth, height=0.1\textwidth]{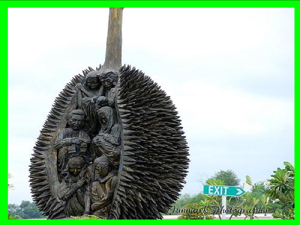} &
\includegraphics[width=0.1\textwidth, height=0.1\textwidth]{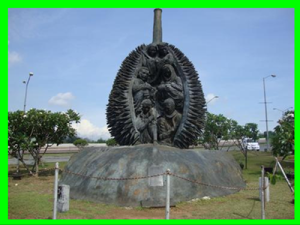} &
\includegraphics[width=0.1\textwidth, height=0.1\textwidth]{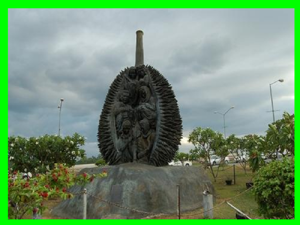}\\
\includegraphics[width=0.1\textwidth, height=0.1\textwidth]{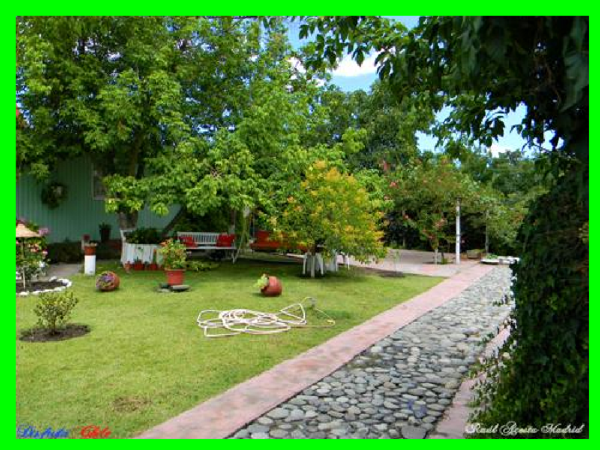} & 
\includegraphics[width=0.1\textwidth, height=0.1\textwidth]{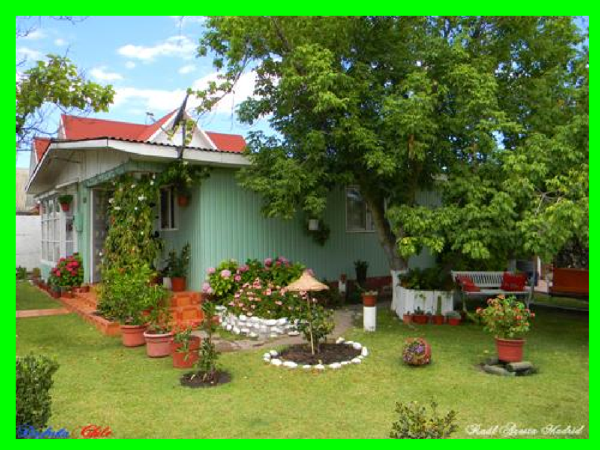} &
\includegraphics[width=0.1\textwidth, height=0.1\textwidth]{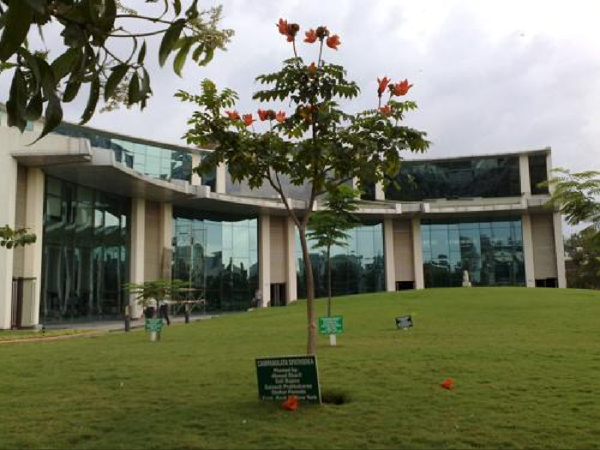} &
\includegraphics[width=0.1\textwidth, height=0.1\textwidth]{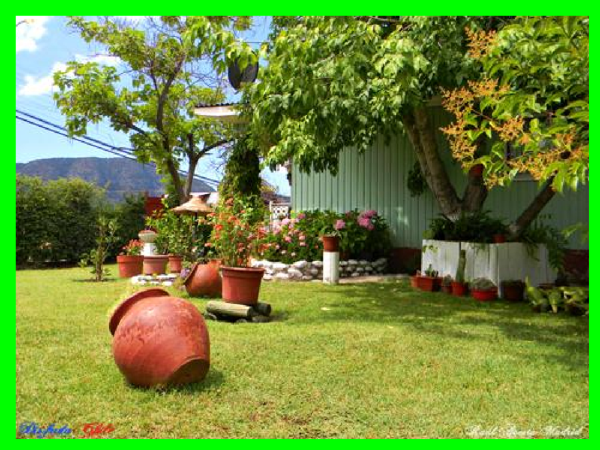}\\
\hline
\end{tabular} &
\begin{tabular}{|@{\hspace{1.5mm}}c@{\hspace{1.5mm}}c@{\hspace{1.5mm}}c@{\hspace{1.5mm}}c@{\hspace{1.5mm}}|}
\hline
\includegraphics[width=0.1\textwidth, height=0.1\textwidth]{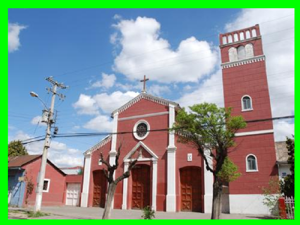} & 
\includegraphics[width=0.1\textwidth, height=0.1\textwidth]{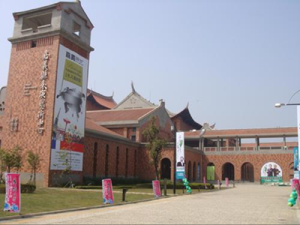} &
\includegraphics[width=0.1\textwidth, height=0.1\textwidth]{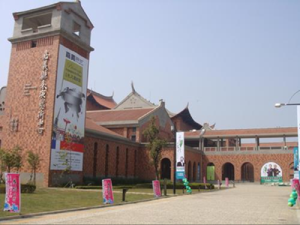} &
\includegraphics[width=0.1\textwidth, height=0.1\textwidth]{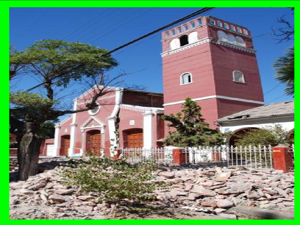}\\
\includegraphics[width=0.1\textwidth, height=0.1\textwidth]{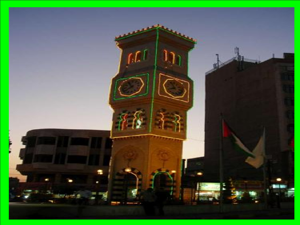} & 
\includegraphics[width=0.1\textwidth, height=0.1\textwidth]{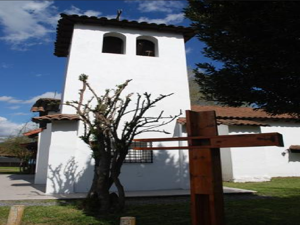} &
\includegraphics[width=0.1\textwidth, height=0.1\textwidth]{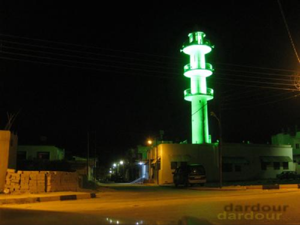} &
\includegraphics[width=0.1\textwidth, height=0.1\textwidth]{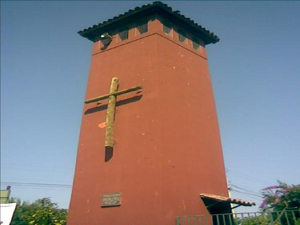}\\
\includegraphics[width=0.1\textwidth, height=0.1\textwidth]{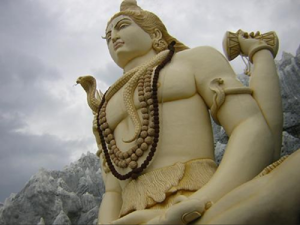} & 
\includegraphics[width=0.1\textwidth, height=0.1\textwidth]{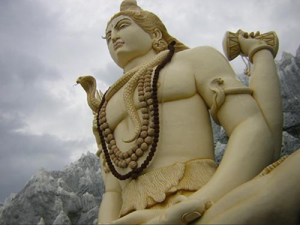} &
\includegraphics[width=0.1\textwidth, height=0.1\textwidth]{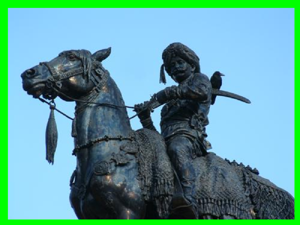} &
\includegraphics[width=0.1\textwidth, height=0.1\textwidth]{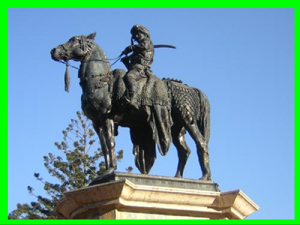}\\
\includegraphics[width=0.1\textwidth, height=0.1\textwidth]{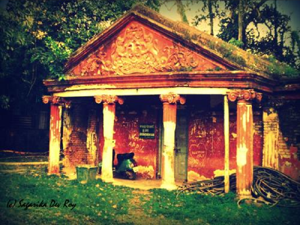} & 
\includegraphics[width=0.1\textwidth, height=0.1\textwidth]{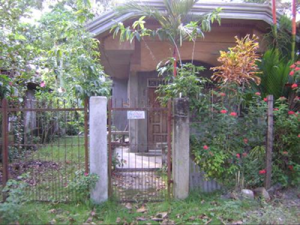} &
\includegraphics[width=0.1\textwidth, height=0.1\textwidth]{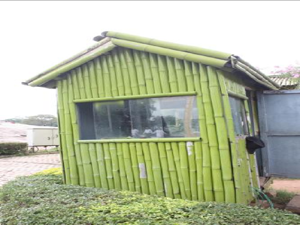} &
\includegraphics[width=0.1\textwidth, height=0.1\textwidth]{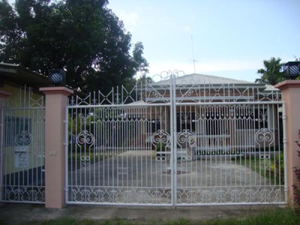}\\
\includegraphics[width=0.1\textwidth, height=0.1\textwidth]{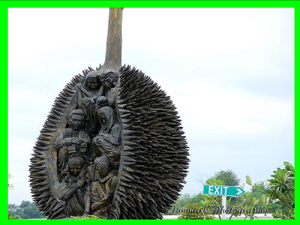} & 
\includegraphics[width=0.1\textwidth, height=0.1\textwidth]{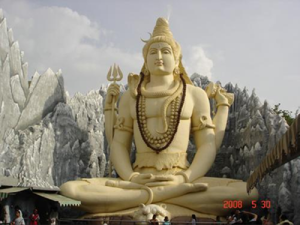} &
\includegraphics[width=0.1\textwidth, height=0.1\textwidth]{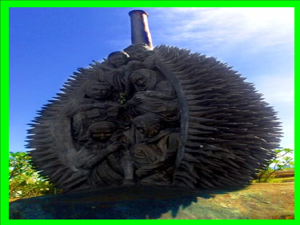} &
\includegraphics[width=0.1\textwidth, height=0.1\textwidth]{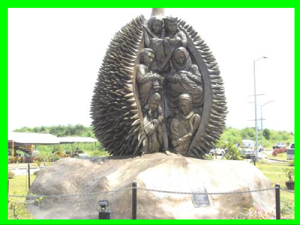}\\
\includegraphics[width=0.1\textwidth, height=0.1\textwidth]{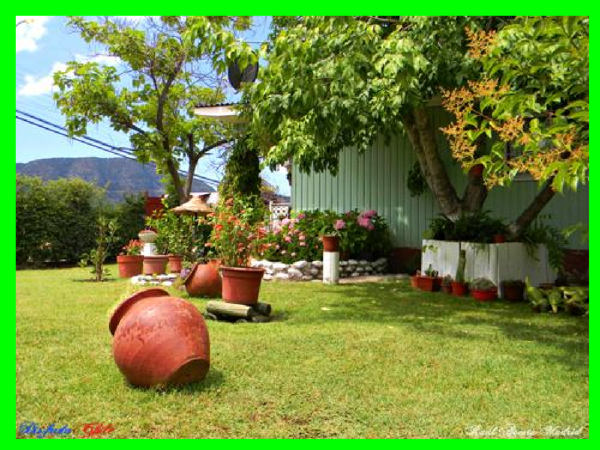} & 
\includegraphics[width=0.1\textwidth, height=0.1\textwidth]{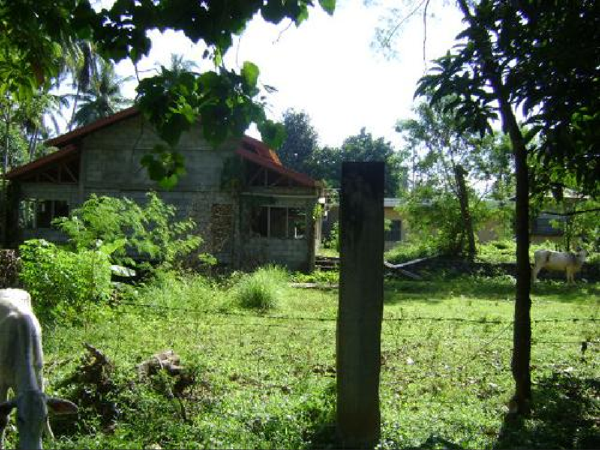} &
\includegraphics[width=0.1\textwidth, height=0.1\textwidth]{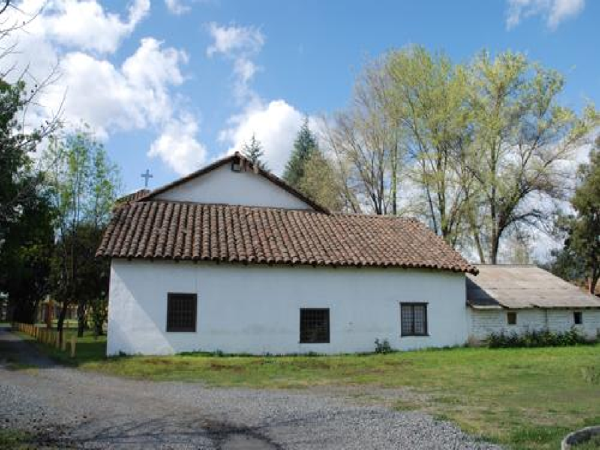} &
\includegraphics[width=0.1\textwidth, height=0.1\textwidth]{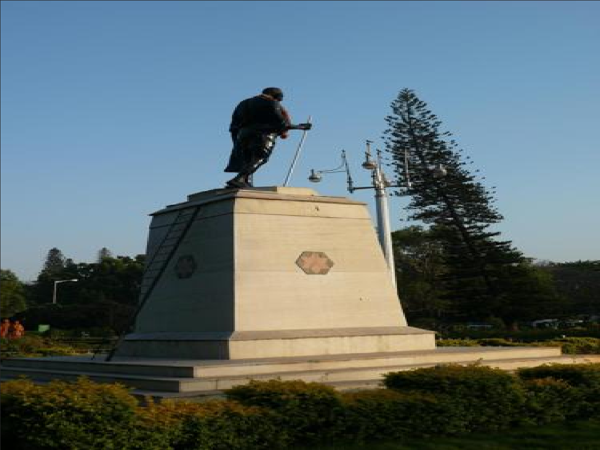}\\
\hline
\end{tabular} \\
Query Image & Average Pooling from pool5 Layer  & FC7 
\end{tabular}
\caption{Qualitative Comparison of the proposed pooling from layer 5 vs using FC7 features on GeoPlaces5K dataset. Images are ranked from left to right. The images which are retrieved correctly are surrounded by green rectangle. PCA and whitening is applied on both of the methods. One interesting observation from these query images is that in the 4th row, all the images being retrieved by fc7 is from the same category (house) but they are not the correct instance. Whereas, pooling from pool5 layer can retrieve images of the same instance. The images are retrieved using {\em Places} pre-trained CNN.}
\label{fig:FC7vsPool5}
\end{figure*}

For image retrieval, images are retrieved according to the cosine distance between the proposed representation of the query image and reference set images. Since convolutional neural networks are not invariant to large rotations, for each image in the reference set we compute the proposed feature representation for 4 different orientations: $0^\circ$, $90^\circ$, $180^\circ$, and $270^\circ$. The distance between query image is defined as the closest distance between the representation of query image and the representation of one of the four rotated images corresponding to each reference image. Figure~\ref{fig:AvgMaxHybridPooling} shows different query images from INRIA Holidays dataset and the top 3 retrieved images using representations with different pooling strategies. As mentioned before, max pooling  is really effective when there is large scale variation between the query image and the reference image. Note that in the last two query images of Figure~\ref{fig:AvgMaxHybridPooling}, hybrid pooling representation is able to retrieve the matching image, while none of the max nor average pooling are able to retrieve the same instances. Figure~\ref{fig:FC7vsPool5} also compares the top retrieved images using fc7 and average pooling on layer pool5.

\section{Experiments}
In the experimental section we evaluate the effectiveness of our representation by comparing the performance of commonly used fc7 features with pool5 layer features on both the image retrieval and the scene categorization tasks. 
The representations are obtained using {\em ImageNet} and {\em Places} networks respectively. 
We examine the effects of the proposed pooling strategies on different datasets.  At last we examine the effectiveness of the proposed representation on a new GeoPlaces5K image retrieval dataset, which contains large variety of scenes with large variations in appearance and viewpoint.

\subsection{Datasets}
We evaluate our approach on the following datasets:
\begin{enumerate}
\item {\bf INRIA Holidays Dataset~\cite{Jegou-ICCV2012}:} This dataset contains 1491 images taken by cellphones at different places and different countries. The images are taken at the same time but with different translation, rotation, and moderate viewpoint changes. There are 500 query images in this dataset and it is evaluated using mean average precision mAP defined in~\cite{Jegou-ICCV2012}.
\item {\bf GeoPlaces5K Dataset :} We obtained this dataset by collecting 5332 images from  5 different countries and 3 different continents. The dataset contains 100 query images and there are 859 images which are matching with the query images. The images are taken at different time of the day (day or night) and from significantly different viewpoints. The distracting images in the dataset are chosen from the locations in vicinity of query images. The minimum distance between each of the distracting images from any of the query images is at most 0.5km. Our results show that this dataset has novel characteristics which are not present in other datasets for image retrieval. Similar to INRIA Holidays dataset, this dataset is evaluated using mAP.
\item {\bf SUN397 Dataset~\cite{SUNCVPR10}:} This dataset contains images from 397 scene categories. There are 10 train/test splits available where each split consist of 50 training images and 50 test images. The evaluation criterion for this dataset is the average classification error on each scene categories over the 10 splits. 
\end{enumerate}
\subsection{Image Retrieval Analysis}
We evaluate our approach using pre-trained convolutional neural network on ImageNet\cite{Krizhevsky-NIPS2012} and Places\cite{cnnPlaces-NIPS2014}. We compare the performance of different pooling methods on both representations. We compared the result of our method with the method of \cite{Lazebnik-arXiv2014}. Table~\ref{tab:Holidays} shows that our method is superior using the same pre-trained CNN. One of the reasons is that our method uses pool5 layer which captures generic semantic concepts which are less dependent on the specific training objective of CNNs. In addition, our feature representation is 48 times smaller which makes it more suitable for the nearest neighbor image retrieval. Lower feature dimensionality has several benefits: 1) the nearest neighbour retrieval \footnote{We use cosine distance in our implementation. However, using Euclidean distance instead of cosine distance does not effect the results on INRIA Holidays dataset.} performs better in lower dimensions; 2) the required space for storing the image representation is much smaller using our method. Another important factor which is also observed in \cite{Lazebnik-arXiv2014} and \cite{Chum-CVPR2012} is applying PCA before whitening. Note that we are not reducing the dimensionality of the features. It is worth mentioning here that whitening is applied on all of the methods in Table\ref{tab:Holidays}. The third row of Table~\ref{tab:Holidays} shows that when using our method on INRIA Holidays dataset, the difference between {\em Places} CNN and {\em ImageNet} CNN is not significant.

\begin{table}
\centering
\caption{Evaluation on the INRIA Holiday Dataset with respect to mAP and feature dimensionality}
\label{tab:Holidays}
\begin{tabular}{|c|c|c|}
\hline
Method & Dim. & mAP \\
\hline
FC7 (Places CNN) & 4096 & 70.24\\
FC7 (ImageNet CNN)& 4096 & 68.30\\
Gong et al. \cite{Lazebnik-arXiv2014} (ImageNet CNN) & 12288 & 80.18\\
\hline
Max pooling (Places CNN)  & 256 & 73.72\\
Max Pooling (ImageNet CNN) & 256 & 70.45\\
Avg Pooling (Places CNN) & 256 & 76.72\\
Avg Pooling (ImageNet CNN) & 256 & 73.21\\
Hybrid Pooling (Places CNN) & 512 & 79.24\\
Hybrid Pooling (ImageNet CNN) & 512 & 76.34\\
\hline
Max pooling + PCA (Places CNN)  & 256 & 77.21\\
Max Pooling + PCA (ImageNet CNN) & 256 & 76.21\\
Avg Pooling + PCA (Places CNN) & 256 & {\bf 82.86}\\
Avg Pooling + PCA (ImageNet CNN) & 256 & {\bf 81.22}\\
Hybrid Pooling + PCA (Places CNN) & 512 & 80.11\\
Hybrid Pooling + PCA (ImageNet CNN) & 512 & 79.39\\
\hline
\end{tabular}
\end{table}

We further investigate the difference between {\em Places} CNN vs {\em ImageNet} CNN derived features on our GeoPlaces5K dataset. This dataset is collected from Panoramio in wild and there is large variation between viewpoint and time of day. This dataset has no overlapping images with {\em Places} nor {\em ImageNet} datasets and has more clutter than INRIA Holidays dataset. Table \ref{tab:Panoramio} shows that using the same method but on the {\em Places} pre-trained CNN leads to better performance. The 6 \% margin between {\em Places} and {\em ImageNet} CNN features on GeoPlaces5K dataset acknowledges the observation in \cite{Zhou-ICLR2014}; Zhou et al. \cite{Zhou-ICLR2014} showed that the pool5 layer of {\em Places} CNN captures more information about discriminant elements of scene categories. Another observation, which is consistent on both INRIA Holidays and our GeoPlaces5k datasets, is that average pooling performs better than max pooling. As mentioned before, average pooling is more robust against various distractors but susceptible to scale change. However, max pooling is more robust to the scale changes. The superiority of average pooling with respect to max pooling could be attributed to the fact that the false positive detections on different feature maps of pool5 layer have more negative impact than sensitivity to the scale change. Hybrid pooling in between of the max pooling and average pooling. Sometimes hybrid pooling even outperform both of the max and average pooling. Table~\ref{tab:Holidays} shows that hybrid pooling performs better than average pooling and max pooling without applying PCA.

\begin{table}
\centering
\caption{Evaluation on GeoPlaces5K dataset using different pre-trained CNNs on ImageNet and Places}
\label{tab:Panoramio}
\begin{tabular}{|c|c|}
\hline
Pooling Method  & mAP \\
\hline
Average Pooling + PCA (ImageNet CNN)& 35.70 \\
Average Pooling + PCA (Places CNN)& {\bf 41.03} \\
\hline
Max Pooling + PCA (ImageNet CNN)& 27.55 \\
Max Pooling + PCA (Places CNN)& {\bf 33.32} \\
\hline
Hybrid Pooling + PCA (ImageNet CNN)& 30.49 \\
Hybrid Pooling + PCA (Places CNN)& {\bf 36.05} \\
\hline
FC7 + PCA (ImageNet CNN) & 29.75\\
FC7 + PCA (Places CNN) & {\bf 36.04}\\
\hline
\end{tabular}
\end{table}
\subsection{Scene Classification Analysis}
We also applied the proposed feature representation to the problem of scene classification. We evaluated the scene classification on the SUN397 dataset.  For each image, features are computed using Caffe \cite{jia2014caffe}. Caffe computes the features over 10 crops ((1 center + 4 corners)* 2 mirrors). For each image, the feature representation for all the 10 crops is stored. The n-way classification is done using JSGD package \cite{akata:hal-00835810} with 100 epochs, regularization factor of $1e-5$, and learning rate of $0.2$. An image is classified as a category if at least 6 crops out of 10 crops are classified as positive for that category. Table~\ref{tab:SUN397} summarizes the results on all 397 scene categories. {\em Places} has better performance due to the fact that the categories in the SUN397 dataset are overlapping with categories with {\em Places} dataset. One interesting trend in Table~\ref{tab:SUN397} is that the classification accuracy increases with the increase in feature dimensionality. Low dimensional feature vector was favorable in image retrieval comparing to \cite{Lazebnik-arXiv2014}. However, more features means higher dimensional space  making the separability between the data points easier to attain. As a result, our method cannot achieve top of the line performance. In 397-way classification, Xiao et al.\cite{SUNCVPR10} achieved 38\% on the whole dataset and 34.5\% on subset of 24 categories. In order to empirically show that our proposed feature dimension is not good enough for large number of classes, we performed the classification on the subset of 24 categories which is mention in \cite{SUNCVPR10}. Using smaller number of categories average pooling from pool5 layer of ImageNet CNN gives 65.92\%. This shows that our current feature representation although suitable for retrieval or small classification problem, it does not perform as well for categorization problems with large number of classes. 
\begin{table}
\centering
\caption{Evaluation on the SUN397 dataset with respect to average precision and feature dimensionality}
\label{tab:SUN397}
\begin{tabular}{|c|c|c|}
\hline
Method & Dim. & mAP \\
\hline
Xiao et al. \cite{SUNCVPR10} & -- & 38.00\\
\hline
Gong et al. \cite{Lazebnik-arXiv2014} (ImageNet CNN) & 12288 & 51.98\\
Donahue et al. \cite{DeCafDonahue-2013} (ImageNet CNN) & 4096 & 40.94\\
\hline
Avg pooling + PCA (Places CNN)  & 256 & 41.031\\
Avg Pooling + PCA (ImageNet CNN) & 256 & 35.70\\
Max Pooling + PCA (Places CNN) & 256 & {33.32}\\
Max Pooling + PCA (ImageNet CNN) & 256 & {27.55}\\
Hybrid Pooling + PCA (Places CNN) & 512 & 51.54\\
Hybrid Pooling + PCA (ImageNet CNN) & 512 & 43.69\\
\hline
\end{tabular}
\end{table}
\section{Discussion}
We proposed simple, yet effective, image representation derived from CNNs pre-trained on {\em ImageNet} and {\em Places} datasets.
Our approach is motivated by recent understanding and visualizations of the semantic information and associated invariances captured by different 
layers of convolutional networks \cite{DarrellSemSeg2014}\cite{ZeilerFergus-ECCV2014}\cite{Zhou-ICLR2014}. 

The feature computation stage of our method is very simple and computationally efficient, which is favorable when scaling to large scale datasets. Unlike other methods where multiple image windows at multiple scales are passed through the network, our method processes image by passing it through the network only once. Instead of aggregating fc7 features at different scales of the image, multi-scale pooling on the pool5 layer can be done without exerting extra computational cost.
The low dimensionality of the proposed feature representation makes it suitable for the image retrieval using the nearest neighbor or approximate nearest neighbor techniques, which suffer more  in higher dimensions. 
The proposed method achieves comparable performance with respect to the state-of-the-art on the scene categorization, but it does not scale well for large number of classes. In such settings higher dimensional feature representations could improve the separability between large number of classes and therefore the classification accuracy.

Our results show that training CNNs on different datasets, while keeping the architecture intact, makes significant difference. We evaluated pre-trained CNNs on {\em Places} and {\em Imagenet} networks and observed, not surprisingly, that the pre-trained {\em Places} network  consistently outperforms the CNN trained on {\em Imagenet}  on both the image retrieval on INRIA Holidays, GeoPlaces5K and the SUN367 scene classification which are all scene datasets. 
This is due to the fact that {\em Places} CNN focuses on detecting discriminative scene elements whereas {\em ImageNet} CNN focuses on object parts. 

The newly introduced  GeoPlaces5K dataset has large variation in the appearance due to images from different continents, different times of day, significant viewpoint change and less usual scenes
compared to INRIA dataset. It also more likely less visual similarity with the images used to train {\em Places} CNN. 
This indicates that the success of repurposing the existing architectures and representations  critically depends on the dataset and characterization of the difference between the source and target datasets as pointed out in~\cite{DeCafDonahue-2013}. 
The performance on the new dataset can be likely further improved by deploying previously suggested fine-tuning strategies.  Another open question 
is the one of the choice of the right loss function for the image retrieval tasks, where the objective is different that the one of categorization.  We will make the dataset available. 
{\small
\bibliographystyle{ieee}
\bibliography{cnn}
}

\end{document}